\documentclass[lettersize,journal]{IEEEtran}
\usepackage{amsmath,amsfonts}
\usepackage{algorithmic}
\usepackage{algorithm}
\usepackage{array}
\usepackage[caption=false,font=normalsize,labelfont=sf,textfont=sf]{subfig}
\usepackage{textcomp}
\usepackage{stfloats}
\usepackage{url}
\usepackage{verbatim}
\usepackage{graphicx}
\usepackage{cite}
\usepackage{hyperref}
\usepackage{makecell}
\usepackage{booktabs}
\usepackage{multirow}
\usepackage{ulem}
\usepackage{caption}
\usepackage{booktabs}
\usepackage{xcolor}
\usepackage{mathrsfs}
\newcommand{\kdboldred}[1]{\textbf{\textcolor{black}{#1}}}

\begin{document}

\title{Hybrid Attention Model Using Feature Decomposition and Knowledge Distillation for Blood Glucose Forecasting}



\author{Ebrahim Farahmand$^*$,
\IEEEmembership{Student Member, IEEE}
Shovito Barua Soumma$^*$, \IEEEmembership{Student Member, IEEE}, 
Nooshin Taheri Chatrudi, \IEEEmembership{Student Member, IEEE}, and 
Hassan Ghasemzadeh, \IEEEmembership{Senior Member, IEEE}

\thanks{Authors are with the College of Health Solutions at Arizona State University, Phoenix, AZ 85004, USA. E-mail: \{efarahma, shovito, ntaheric, hassan.ghasemzadeh\}@asu.edu}
\thanks{*Both authors contributed equally to this research.}
}

\markboth{IEEE Transactions on Mobile Computing,~Vol.~14, No.~8, August~2026}%
{Shell \MakeLowercase{\textit{et al.}}: A Sample Article Using IEEEtran.cls for IEEE Journals}


\maketitle

\begin{abstract}

The availability of continuous glucose monitors (CGMs) as over-the-counter commodities \textcolor{black}{has} created a unique opportunity to monitor a person's blood glucose levels, forecast blood glucose trajectories, and provide automated interventions to prevent devastating chronic complications that arise from poor glucose control. However, forecasting blood glucose levels (BGL) is challenging because blood glucose changes consistently in response to food intake, medication intake, physical activity, sleep, and stress. It is particularly difficult to accurately predict BGL from multimodal and irregularly sampled mobile sensor data and over long prediction horizons. Furthermore, these forecasting models need to operate in real-time on edge devices to provide in-the-moment interventions. To address these challenges, we propose \textbf{GlucoNet}\footnote{Code base is available at: \href{https://github.com/shovito66/GlucoNet}{\textcolor{blue}{https://github.com/shovito66/GlucoNet}}}, 
an AI model to forecast blood glucose patterns using sensor data about behavioral and physiological health. GlucoNet devises a feature decomposition-based lightweight transformer model that incorporates patients' behavioral and physiological data (e.g., blood glucose, diet, medication) and transforms sparse and irregular patient data (e.g., diet and medication intake data) into continuous features using a mathematical model, facilitating better integration with the BGL signals. Given the non-linear and non-stationary nature of blood glucose signals, we propose a decomposition method to extract both low-frequency (long-term) and high-frequency (short-term) components from the BGL signals, thus enabling the model to capture complex glucose dynamics for accurate forecasting. To reduce the computational complexity of transformer-based predictions, we propose to employ knowledge distillation (KD) to compress the transformer model. 
\color{black}
Our comprehensive analysis on two real-world T1D cohorts demonstrates that GlucoNet achieves a 35\% improvement in RMSE, a 33\% improvement in MAE, and a 62\% reduction in the number of parameters over state of the art work such as PatchTST on the OhioT1DM dataset (12 patients), while additional experiments on the AZT1D dataset (25 patients), together with extensive ablation and robustness analyses, further demonstrate its generalizability and stability.
These results underscore GlucoNet's potential as a compact and reliable tool for real-world diabetes prevention and management.
\end{abstract}


\section{Introduction}
Glucose control (i.e., maintaining blood glucose within a normal range such as 70--180 mg/dL) is not only central to managing diabetes but also important in preventing cardiovascular disease, stroke, and cancer. Additionally, glucose control has been shown to be effective in losing weight, optimizing mental health, suppressing food cravings, and improving sleep. Sensor technologies that can monitor a person's blood glucose levels can play a crucial role in glucose control. Until recently, the use of continuous glucose monitoring (CGM) was largely confined to individuals with diabetes and was primarily prescribed for clinical disease management. However, CGM technologies are rapidly becoming more accessible, with several devices now available directly to consumers as over-the-counter products. This increased availability has expanded the potential role of CGMs beyond traditional diabetes care to broader applications in health monitoring, lifestyle optimization, and early risk detection related to impaired glucose regulation. As part of the broader ecosystem of remote health monitoring and human-centered IoT systems, CGM has emerged as a practical and increasingly scalable modality for continuous metabolic assessment and personalized disease management~\cite{rokni2016plug,soumma2024ssl}. These advances create new opportunities for data-driven behavioral and clinical interventions aimed at improving glycemic control and reducing the risk of downstream physical and mental health complications associated with dysglycemia.


Forecasting blood glucose levels based on factors such as diet, physical activity, and medication will enable patients and providers to take preventive actions against abnormal glucose events such as hyperglycemia and hypoglycemia. Hyperglycemia, characterized by blood glucose levels (BGL) exceeding $180 mg/dL$, can lead to complications affecting the cardiovascular system, eyes, kidneys, and nerves. On the other hand, precise medication dosing (e.g., insulin in patients with diabetes) is crucial, as excessive amounts can lead to hypoglycemia. A blood glucose level (BGL) below  $70 mg/dL$, defined as Hypoglycemia, is a dangerous condition that can lead to fainting, coma, or even death in severe cases~\cite{shuvo2023deep}. Therefore, accurately predicting future glucose levels helps determine optimal insulin dosages, plan meals, and organize exercise regimens. This allows for proactive measures to reduce the risk of dangerous glycemic events.

Traditional machine learning algorithms such as ARIMA models~\cite{shanthi2011novel} and Random Forests~\cite{georga2015evaluation} have investigated methods for predicting BGLs. However, deep learning models have demonstrated significant advantages over traditional machine learning algorithms in classification~\cite{10552876} and prediction tasks~\cite{armandpour2021deep} by effectively addressing complex, nonlinear problems. In recent years, these models have shown high-performance levels in blood glucose forecasting, demonstrating the potential to improve diabetes management and reduce long-term complications associated with the condition. However, despite all these advancements in technologies and prediction modeling, many patients still experience abnormal glucose events (hyper and hypo events). Therefore, a more accurate modeling is needed.

\textcolor{black}{BGL fluctuations arise from multiple physiological and behavioral factors, including carbohydrate intake, insulin administration, mental stress, sleep quality, and physical activity. Therefore, relying solely on CGM measurements may not provide sufficient information to fully capture and model these complex dynamics. Blood glucose regulation is shaped by interacting processes such as carbohydrate absorption, insulin kinetics, physical exertion, psychological stress, and circadian influences. These variables contribute to both gradual metabolic trends and rapid transient fluctuations that cannot be inferred from glucose measurements alone. Moreover, CGM sensors measure glucose in interstitial fluid rather than blood, reflecting the downstream outcome of these processes rather than their initiating mechanisms. Consequently, CGM readings typically lag true blood glucose changes by approximately 5–10 minutes and exhibit measurement variability, with reported MARD values commonly in the range of 9–12\% for commercial systems~\cite{danne2017international}. When a forecasting model relies exclusively on CGM data, it must implicitly approximate the effects of these unobserved physiological drivers, often reducing prediction accuracy during rapid post-prandial or post-insulin excursions. Incorporating multimodal features such as carbohydrate intake and insulin dosing thus provides essential causal information that complements CGM data and enables a more comprehensive representation of glucose dynamics.} 

Some researchers identified that the combination of CGM data with some of these variables improves the performance of the forecasting blood glucose model~\cite{shuvo2023deep}. However, these deep learning models cannot provide acceptable accuracy for forecasting due to the high fluctuations in blood glucose time series data. Moreover, these models can be affected by data distribution irregularities and noise, potentially distorting outcomes~\cite{xue2024bgformer}. To address these issues, recent researchers used the transformer model, which utilizes self-attention mechanisms to forecast BGL~\cite{chen2024multi}. Transformer is a powerful deep learning model that has the ability to simultaneously process information across the entire input sequence and effectively capture temporal dependencies~\cite{xue2024bgformer}. However, implementing transformers in wearable devices \textcolor{black}{pose} challenges due to their high computational demands and complex configuration requirements.

Predicting BGL directly from raw CGM data can reduce forecasting accuracy due to the BGL’s non-linear and non-stationary patterns. Recent research indicates that employing multi-scale decomposition techniques can effectively mitigate the impact of non-stationary on prediction outcomes~\cite{ur2019multivariate,yousefi2023short,kaur2021eeg}. The decomposition method breaks down the complex time series into more manageable components, potentially improving the overall accuracy of forecasting models~\cite{wang2020blood}. 

\begin{figure}[t]
    \centering
    \includegraphics[width=1\linewidth]{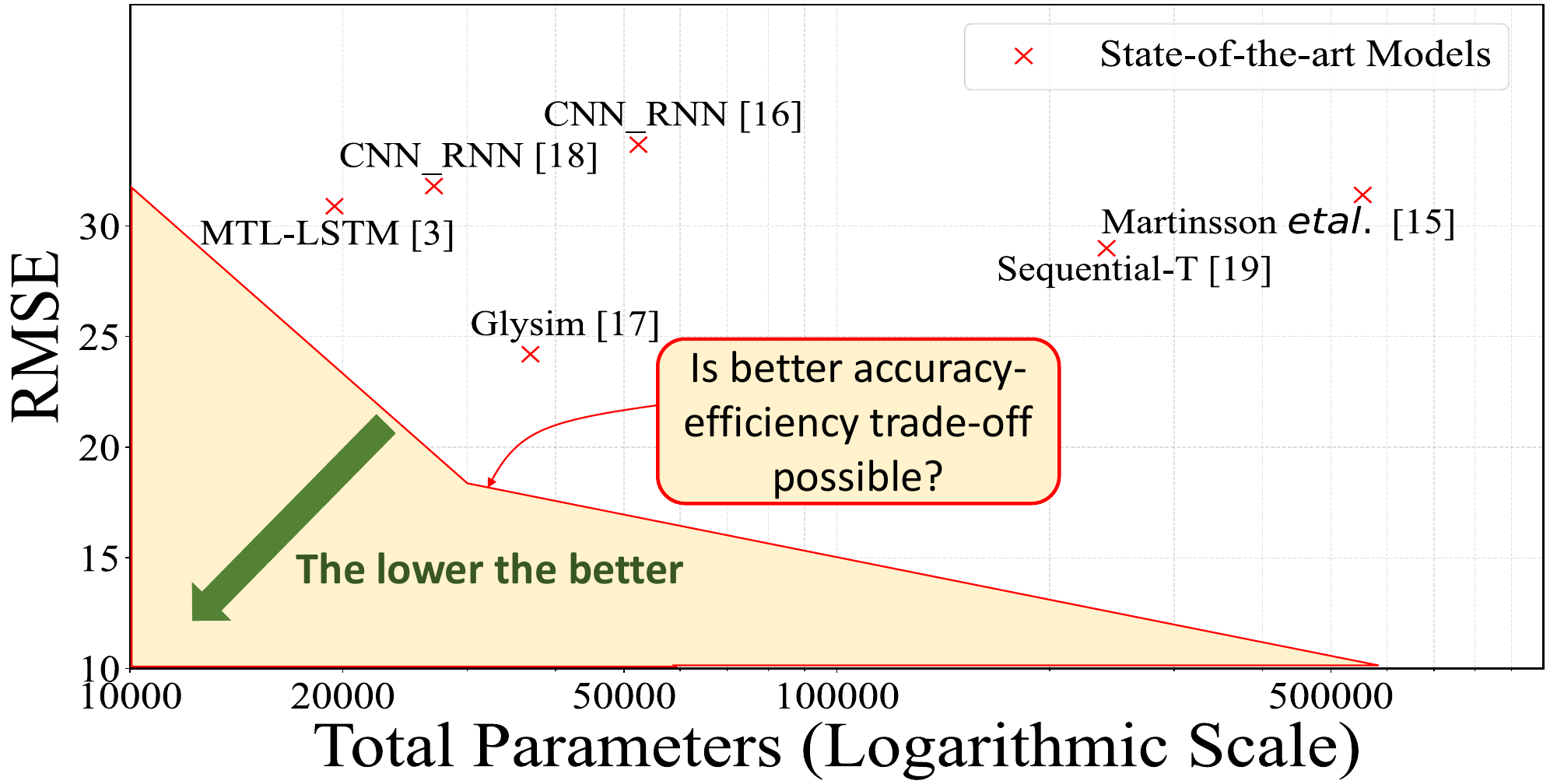}
    \caption{Accuracy-efficiency trade-offs MTL-LSTM~\cite{shuvo2023deep}, Martinsson \textit{et al.}~\cite{martinsson2020blood}, 
CNN-RNN~\cite{freiburghaus2020deep}, GlySim~\cite{arefeen2023glysim}, CNN-RNN~\cite{daniels2021multitask}, and Sequential-T~\cite{11250578}.  RMSE is used to quantify the accuracy (quality) of the models, and the total parameters of the model are used to quantify the efficiency (RMSE vs. Total parameters).}
    \label{fig:limitation}
\end{figure}

\textbf{Key Limitations and Associated Challenges:}
The following points highlight the key limitations of state-of-the-art works and also present the associated challenges blood glucose prediction faces.
\begin{itemize}
    \item Unable to accurately predict abnormal events, such as hyperglycemia and hypoglycemia, which are critical in blood glucose prediction.
    \item Blood glucose levels are influenced by numerous variables, including carbohydrate intake, insulin dosage, stress levels, sleep patterns, and physical activity. Notably, these data are generated at varying rates and sampling frequencies, creating significant sampling irregularities across the multi-modal inputs.
    \item Blood glucose time series data exhibits strong time-varying characteristics, with non-linear and non-stationary properties that make direct prediction challenging using raw CGM data.
\end{itemize}
Some recent works using modern deep learning models aim to address the limitation of forecasting blood glucose. These models include Martinsson \textit{et al.}~\cite{martinsson2020blood}, 
CNN-RNN~\cite{freiburghaus2020deep}, GlySim~\cite{arefeen2023glysim}, CNN-RNN~\cite{daniels2021multitask}, and MTL-LSTM~\cite{shuvo2023deep}. Although these models offer forecasting models for blood glucose, their efficiency in terms of total parameters drastically increases as accuracy requirements increase, which negatively impacts resource demand. 
Fig~\ref{fig:limitation} presents the design space of the accuracy-efficiency of these models. 
The figure highlights that as the accuracy requirements increase, the cost of the optimal design in terms of total parameters increases drastically. 
\textit{Thus, a novel sophisticated forecasting model is required that can offer improved accuracy-efficiency trade-offs.} 

\begin{table*}[t]
    \centering
    \begin{tabular}{p{1.2cm}|p{1.9cm}|p{2.2cm}|p{5cm}|p{5.8cm}}
    \toprule
        \textbf{Study} &\textbf{Method}&\textbf{Input Features} &\textbf{Technique}&\textbf{Attributes and Limitation}\\
        \midrule
         
         ~\cite{wenbo2021blood, wang2020blood} & Deep Learning (LSTM) & BGL and Intrinsic Mode Functions (IMFs) &Decomposes BG using VMD into IMFs for enhanced predictive accuracy& Large number of LSTM memory cells, Not considering sampling irregularities for input features  \\
         \hline
         
         ~\cite{dragomiretskiy2013variational}&Deep Learning (LSTM) & BGL & Adaptive, noise-resilient decomposition of non-linear signals using VMD & Computationally intensive, Not Suitable for potential edge deployment, Not considering the effect of other features on BGL \\
         \hline
         
         ~\cite{jaloli2023long} &Deep Learning (CNN, LSTM)  & BGL, Carbs, insulin &Using the deep learning model and combined data to capture spatial and temporal glucose patterns & Computationally intensive, Not considering sampling irregularities for input features\\
         \hline
         
         ~\cite{li2019convolutional}& Deep Learning (CNN, LSTM) & BGL & Integrates CNN (high-frequency) and LSTM (long-term dependencies) & Not considering the effect of other features on BGL, Requires high computational resources\\
         \midrule
         \makecell{GlucoNet}& Deep Learning (CNN, LSTM, and Transformer) & BGL, Carbs, insulin & Decomposes BGL using VMD into high and low frequency and integrate 1D CNN, LSTM, and Transformer for glucose forecasting &Transform input features for regulating sampling rate, Using Knowledge Distillation to compress models, Suitable for potential edge deployment, Accuracy-Efficiency tradeoffs \\
         \bottomrule
    \end{tabular}
    \caption{A summary of some works and their comparison with GlucoNet}
    \label{tab:related}
\end{table*}

To address these challenges, we proposed GlucoNet, a multimodal attention model that outperforms state-of-the-art methods in forecasting accuracy over longer prediction horizons. The proposed model incorporates LSTM and transformer models to forecast future blood glucose levels in diabetes patients, utilizing data from continuous glucose monitoring (CGM) devices, carbohydrate intake, and insulin administration. The framework extends Variational Mode Decomposition (VMD) to decompose BGL into low-fluctuation and high-fluctuation signal modes. The LSTM is used to forecast low-fluctuation signals. Furthermore, the Transformer's attention mechanism is specifically employed to predict high-fluctuation signals. Finally, we used the knowledge distillation (KD) method ~\cite{hinton2015distilling} to reduce the complexity and size of the transformer model. Knowledge distillation is a compression technique that is drawn from the transfer of knowledge between teachers and students. 


\textcolor{black}{To the best of our knowledge, GlucoNet is the first framework evaluated on the OhioT1DM dataset that reconstructs behavioral events into continuous physiological features to mitigate sampling irregularities and applies time-series decomposition for blood glucose forecasting. This study introduces GlucoNet, a hybrid framework that integrates several new design components to improve blood glucose forecasting accuracy and efficiency. The model converts sparse behavioral inputs into continuous physiological signals, decomposes complex glucose dynamics using VMD, and leverages a parallel LSTM–Transformer hybrid architecture for accurate blood glucose prediction. Furthermore, a knowledge distillation approach is applied to compress the Transformer, achieving lower complexity while preserving accuracy. The combined effect of these innovations leads to a more computationally efficient framework that surpasses existing prediction models on the OhioT1DM dataset.}

\textcolor{black}{The main contributions of this work are summarized as follows:}
\begin{itemize}
\item \textcolor{black}{\textbf{Sparse-to-continuous feature transformation:} Meal and insulin intake events are converted into continuous time-series signals to address sampling irregularities.
\item \textbf{VMD-based signal decomposition:} Variational Mode Decomposition is used to separate low- and high-frequency glucose components, improving stability and prediction accuracy.
\item \textbf{Parallel hybrid forecasting:} A combined LSTM–Transformer model predicts long-term trends and short-term fluctuations in parallel for enhanced adaptability.
\item \textbf{Knowledge distillation for model compression:} A student–teacher KD framework reduces the Transformer’s parameter count while maintaining accuracy.
\item \textbf{Performance improvement:} GlucoNet achieves superior accuracy and efficiency compared to state-of-the-art approaches on the OhioT1DM dataset.}
\end{itemize}

The remainder of this paper is organized as follows: Section~\ref{sec:related_works} reviews related work, Section~\ref{sec:GlucoNet_architecture} presents the GlucoNet architecture, Section~\ref{sec:Experimental_Evaluation} describes the experimental setup, reports the results and analyses, and Section~\ref{sec:conclusion} concludes the paper and outlines future directions.

\section{Related Work}
\label{sec:related_works}
In recent years, blood glucose (BG) forecasting research has focused on addressing the challenges of non-linear, non-stationary time series data in diabetes management. This section reviews relevant approaches, the use of deep learning models, focusing on the application of variational mode decomposition (VMD) in BG prediction, and efforts to improve prediction accuracy and robustness against sensor noise and modeling limitations.

A deep learning model that has recently been used to predict BGLs is Recurrent Neural Networks (RNNs), which incorporate temporal sequencing into their architectural design, enhancing their capability to analyze time series data~\cite{martinsson2020blood}. Long Short-Term Memory (LSTM) model refined the network unit structure of RNNs~\cite{hochreiter1997long}. The LSTM architecture addresses several limitations of traditional RNNs, such as the vanishing and exploding gradient problems and its challenge in retaining long-term information. The LSTM design can extract features from extended temporal sequences. Therefore, the LSTM model has been successfully applied to forecast blood glucose levels. An LSTM model is proposed in~\cite{van2021machine} by using multiple data variables such as CGM and accelerometer signals to forecast blood glucose levels. The method applied a moving average on blood glucose and accelerometer signals to preprocess them and then \textcolor{black}{pass} these data to LSTM to predict the blood glucose levels. Asiful \textit{et al.} presented a stacked 1DCNN-LSTM model called GlySim to forecast blood glucose levels based on multi-modal data from continuous glucose monitors, insulin pumps, and carbohydrate intake~\cite{arefeen2023glysim}.

Variational Mode Decomposition (VMD) has been widely applied in time series analysis to enhance predictive accuracy by decomposing complex signals into distinct frequency components, known as intrinsic mode functions (IMFs). Proposed by Dragomiretskiy and Zosso, VMD offers a more adaptive and noise-resilient alternative to earlier methods like Empirical Mode Decomposition (EMD)~\cite{dragomiretskiy2013variational}. In blood glucose (BG) prediction, VMD has been used to preprocess signals and address non-stationarity. For instance, Wang et al. introduced a VMD-KELM-AdaBoost model, which first decomposes BG signals into IMFs and then predicts each IMF using a kernel extreme learning machine combined with AdaBoost, achieving improved prediction accuracy across various time horizons~\cite{wang2020blood, wenbo2021blood}. 

Hybrid neural networks, especially those combining convolutional and recurrent layers, have shown promising results in BG prediction by capturing complex spatial and temporal patterns. In ~\cite{jaloli2023long}, authors proposed a CNN-LSTM hybrid model that uses historical glucose data, meal intake, and insulin information, achieving accurate forecasts across multiple horizons. Similarly, Li et al. developed a Convolutional Recurrent Neural Network (CRNN) that integrates convolutional layers to capture high-frequency variations and recurrent LSTM layers for long-term dependencies, yielding high accuracy in both simulated and real datasets~\cite{li2019convolutional}. However, these hybrid architectures are often computationally demanding and less interpretable, limiting their practicality for integration with multiple data sources.
\begin{figure*}[t]
    \centering
    \includegraphics[width=0.9\linewidth]{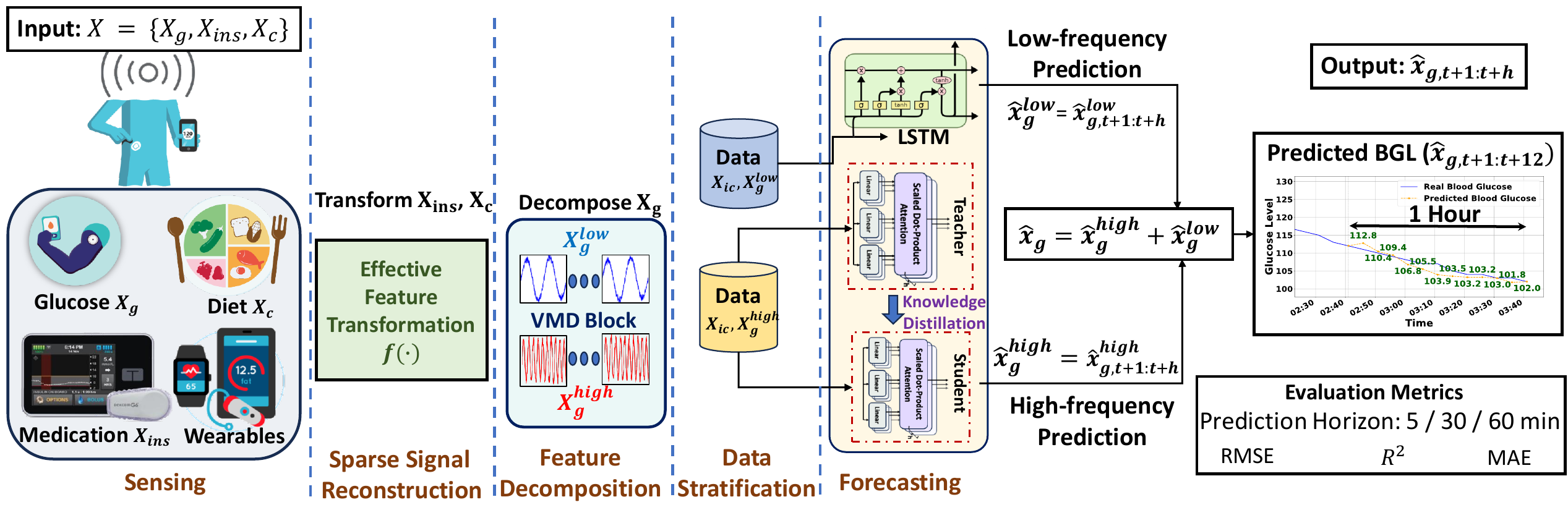}
    \caption{An Overview of GlucoNet includes sensing to measure variables, Sparse signal construction to extract the effective variables, Feature decomposition using the VMD method, Data stratification to combine various variables, and a forecasting module using LSTM and Knowledge Distillation Framework.}
    \label{fig:proposed_method}
\end{figure*}
To improve accuracy and reliability, several studies have introduced preprocessing or model refinements to counter sensor noise and enhance robustness. Rabby et al. implemented a stacked LSTM with Kalman smoothing to correct CGM sensor errors, thereby enhancing accuracy even with noisy data~\cite{rabby2021stacked}. Lee \textit{et al.} highlighted the limitations of conventional evaluation metrics in BG forecasting, especially in closed-loop systems where accuracy metrics may not align with effective control performance. They stressed the need for alternative evaluation methods beyond standard error metrics to better gauge model effectiveness~\cite{lee2024shortcomings}. While these enhancements improve forecasting, they also increase preprocessing requirements and computational costs, which can hinder real-time deployment on resource-limited devices~\cite{mamun2022designing,mamun2022multimodal}.

Knowledge Distillation, implemented by Hinton \textit{et al.}~\cite{hinton2015distilling}, is a compression technique drawn from the knowledge transfer between teachers and students. The main idea is to assist the small student models in emulating the behavior of massive and over-parameterized teacher models. This approach enhances the training of student models and achieves significant compression and acceleration, enabling the student model to occasionally surpass the teacher model in effectiveness.
Self-distillation, introduced in~\cite{zhang2019your}, allows a single neural network to act as both teacher and student, where deeper layers guide the learning of shallower ones. This method reduces model size and computational demands without sacrificing accuracy, making it ideal for resource-limited environments.

In blood glucose level prediction, knowledge distillation has been effectively employed to transfer learned representations from large datasets to smaller, curated datasets. Hameed \textit{et~al}.~\cite{hameed2020investigating} demonstrated that pre-training a teacher model on a large, noisy dataset and applying it to a smaller dataset can improve multi-step prediction accuracy, though direct training on the target dataset often yields better single-step performance. This approach complements progressive self-distillation techniques by enhancing prediction accuracy without increasing model size, making it suitable for resource-constrained environments.

Despite the advancements, these approaches face key limitations: VMD models struggle with computational demands, hybrid neural networks require high resource use, and enhanced methods depend on extensive preprocessing. To address these gaps, this study integrates VMD with a streamlined neural network model that balances accuracy, efficiency, and interpretability, aiming for practical deployment in diabetes management settings.

A summary of the related works and their comparison with the contribution of this paper, i.e., GlucoNet, is presented in Table~\ref{tab:related}.

\section{GlucoNet Architecture}
\label{sec:GlucoNet_architecture}
 GlucoNet model framework for glucose prediction is explained in this section. An overview of our proposed method is presented in Fig.~\ref{fig:proposed_method}. Our proposed GlucoNet framework is made up of \textit{a)} sensing module, \textit{b)} Sparse signal construction module, which is used for extracting the effective carbohydrate intake (Carb) and insulin values, \textit{c)} Feature decomposition module to break down the time series data into several physically meaningful modes (features). \textit{d)} Data stratification module, and \textit{e)} forecasting module. An overview of our proposed method is presented in Fig.~\ref{fig:proposed_method}.  We adopted LSTM and Transformer models to forecast the future blood glucose levels of diabetic patients using BGL data, carbohydrate intake (Carbs), and insulin dosage. We extended the Variational Mode Decomposition (VMD) to decompose the BGLs and divided them into low-fluctuated signals and high-fluctuated signals. The attention mechanism in the Transformer will help predict the high-fluctuated signals. Furthermore, to decrease the complexity and compression of the Transformer model, we developed the knowledge distillation (KD) method. The effectiveness of GlucoNet is demonstrated using the publicly available OhioT1DM dataset~\cite{marling2020ohiot1dm}. We will explain each module in detail in the following sections.

\subsection{Sensing Module}

The task of blood glucose forecasting with multi-modal input data can be formalized as a time series prediction problem. Let $X = \{ \mathbf{x}_{0,0:t}, \mathbf{x}_{1,0:t} \ldots, \mathbf{x}_{n,0:t}\}$ represent the set of $n$ feature/sensor observations in the Sensing module. The $j^{th}$ feature observation $\mathbf{x}_{j,0:t} = \{x_{j,0}, x_{j,1},\ldots, x_{j,t}\}$ where $x_{j,k}\in \mathbb{R}^t$ represent the feature vector of $j^{th}$ sensor's output up to time $t$. In~\cite{shuvo2023deep}, it is shown that BGL measurements from CGM devices, along with insulin and carbohydrate (Carbs) intake, have a significant impact on blood glucose forecasting performance. However, feature ablation analyses indicate that incorporating additional physiological signals, such as skin temperature and heart rate, does not consistently improve prediction accuracy and may introduce variability that limits generalization across different conditions. Therefore, Our proposed method (GlucoNet) aims to utilize combinations of these measurements, i.e. BGL data, carbohydrate (Carbs) intake, and insulin dosage, to develop a forecasting model for blood glucose prediction. 
The corresponding output time series for multi-output forecasting is denoted as $\mathbf{\hat x}_{g,t+1:t+h} = \{x_{g,t+1}, x_{g,t+2}, \ldots, x_{g,t+h}\}$, assuming $g^{th}$ sensor is CGM, which represents multiple future blood glucose \textcolor{black}{level} values across a given prediction horizon (PH). The prediction horizon refers to the length of time into the future for which predictions are made based on past data. In our evaluation, we test our approach using a prediction horizon of $5$, $30$, and $60$ minutes. Given that CGM data are typically recorded at $5$-minute intervals, a PH of $5$ minutes corresponds to $h=1$ samples, a PH of $30$ minutes corresponds to $h=6$ samples, while a PH of $60$ minutes corresponds to $h=12$ samples.

To ensure consistent evaluation across both output settings, we calculate the error metrics, such as root mean square error (RMSE), by comparing the actual future glucose level with the predicted values in the estimated multi-output sequence. Mathematically, we can express the forecasting task as $\mathbf{\hat x}_{g,t+1:t+h} = f(X,\Theta)$,
\noindent where $f(\cdot)$, $\mathbf{\hat{x}}_{g,t+1:t+h}$, and $\Theta$ represent the forecasting model, predicted future glucose values, and model trainable parameters, respectively.



\begin{figure*}[]
    \centering
    \includegraphics[width=0.8\linewidth]{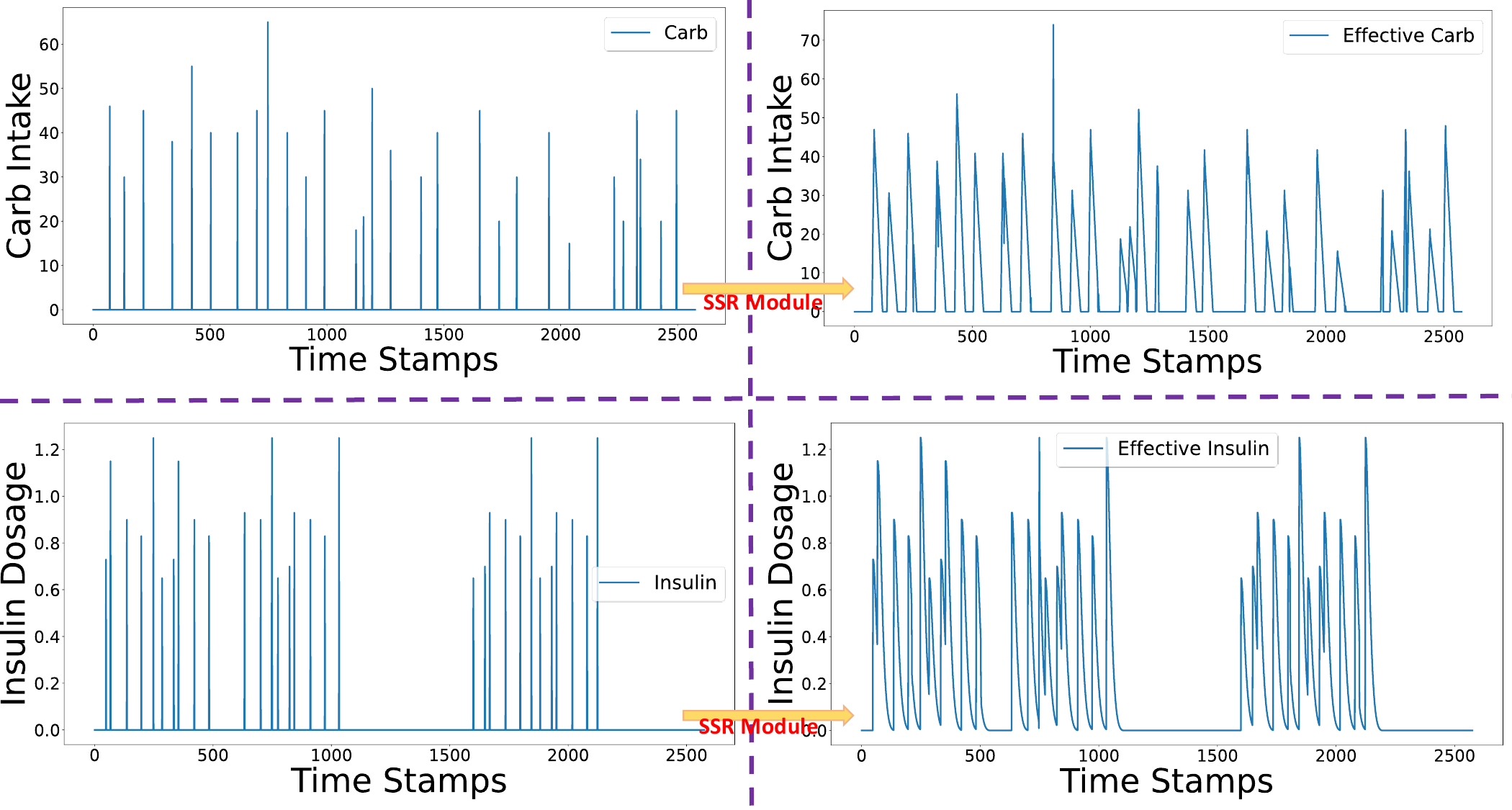}
    \caption{An example of transforming sparse events (carb intake and insulin dosage) using Sparse Signal Reconstruction (SSR) module.}
    \label{fig:effective_event}
\end{figure*}

\subsection{Sparse Signal Reconstruction (SSR) Module
}
\label{Mathematical}
To improve the accuracy of our proposed forecasting of blood glucose, we combined raw CGM values, which are time-series features, with event-based features (e.g., carbohydrate and insulin intake). We proposed transforming event-based features into continuous time-series data. This transformation aims to better represent the physiological effects of carbohydrate intake and insulin on blood glucose levels over time and decrease the effect of sampling irregularities of these data. 

We transformed the sparse meal events into a continuous "operative carbohydrate" feature that represents the ongoing effect of the intake of carbohydrates on blood glucose~\cite{kraegen1981timing}. The carbohydrate intake is transformed from an event-based feature to continuous time series data by Eq.~\ref{eq:meal}. The transformation assumes that carbohydrates begin affecting blood glucose $15$ minutes after the meal time ($t_m$), peak ($t_p$) at $60$ minutes, and gradually decrease over $3$ hours. Note that the time index in Eq.~\ref{eq:meal} is defined with a sampling interval of $5$ minutes. Accordingly, each unit increment in $t$ corresponds to $5$ minutes in real time. For example, a meal occurring at index $t_m$ corresponds to an actual time of $5 \times t_m$ minutes.

\begin{equation}
\hspace{-0.4em} 
\small
C_{op}(t_s) = 
\begin{cases}
0, & 0 \leq t_s-t_{m} \leq 2, \\[8pt]
\frac{(t_s - t_{m+2})}{(t_s - t_{s-1})} \times \alpha_{inc} \times C_{m}, & 3 \leq t_s-t_{m} < 12, \\[8pt]
\left(1 - \frac{(t_s - t_{p})}{(t_s - t_{s-1})} \times \alpha_{dec}\right) \times C_{m}, & 12 \leq t_s-t_{m} < 48
\end{cases}
\label{eq:meal}
\end{equation}

Here, $t_s$, $t_{m}$, and $t_{p}$ present the sampling time, time when the meal is taken, and time when $C_{op}$ reaches its maximum value, respectively. Note that the unit of time is minutes in this equation. $C_{op}$ is the effective carbohydrates at any given time, and $C_{m}$ is the total amount of carbohydrates in the meal. Based on the~\cite{kraegen1981timing}, $\alpha_{inc}$ is the increasing rate, which is set to $0.11$, and $\alpha_{dec}$ is the decreasing rate set to $0.028$.

The other variable significantly impacting blood glucose levels is insulin dosage. Similar to meal carbohydrates, we transform insulin dosage into a continuous active insulin feature. This transformation extended from~\cite{boiroux2010optimal}, based on the insulin activity curve. This transformation accounts for insulin activity's gradual onset, peak, and decay, providing a more accurate representation of insulin's effect on blood glucose over time. The active insulin at any given time is modeled using Eq.~\ref{eq:insulin}. 


\begin{equation}
\small
\begin{aligned}
IOB(t_s) &= 1 - S \times (1-a) \times \bigg\{ \bigg( \frac{t_s^2}{\tau \times t_d \times (1-a)} 
- \frac{t_s}{\tau} - 1 \bigg) \times \\
&\quad\quad e^{-\frac{t_s}{\tau}} + 1 \bigg\}
\end{aligned}
\label{eq:insulin}
\end{equation}

\noindent Where $t_s$ is the sampling time, $t_d$ is the total duration of insulin activity, $\tau$ is the time constant of exponential decay, $a$ is the rise time factor, $S$ is the auxiliary scale factor. The other terms of Eq.~\ref{eq:insulin} are calculated as follows:

\begin{equation}
\tau = t_p \times \left(1 - \frac{t_p}{t_d}\right) / \left(1 - 2 \times \frac{t_p}{t_d}\right)
\end{equation}

\begin{equation}
a = 2 \times \tau / t_d
\end{equation}

\begin{equation}
S = 1 / \left(1 - a + (1 + a) \times e^{-\frac{t_d}{\tau}}\right)
\end{equation}

\textcolor{black}{
In this work, we deliberately use a lightweight SSR-based model for IOB/COB rather than more complex PK/PD formulations. OhioT1DM and AZT1D provide insulin and meal logs at coarse resolution and lack key physiological covariates (e.g., insulin formulation, gastric emptying), making higher‑order PK/PD models hard to identify and prone to overfitting. The SSR kernels instead encode only the well‑known delayed onset, peak, and decay of insulin and carbohydrate action, yielding smooth, robust inputs that match the data granularity while keeping preprocessing cost low enough for edge deployment.
}

The SSR transformation is further motivated by the requirements of time-series forecasting and real-time deployment. In CGM datasets, insulin and meal inputs are sparse events, whereas their physiological effects evolve continuously. Direct use of raw events fails to capture this temporal influence and may result in information loss when past events fall outside the finite input window despite still affecting future glucose levels. The proposed transformation mitigates this by encoding delayed onset, peak response, and gradual decay into continuous signals, enabling more effective learning of glucose dynamics. While more advanced preprocessing approaches (e.g., PK/PD or neural ODE-based models) offer higher physiological fidelity, they require additional parameters and computational resources that are not available in typical wearable datasets and are less suitable for real-time edge deployment.

Fig.~\ref{fig:effective_event} shows an example of transforming the event-based features (e.g., Carb intake and insulin dosage) to the continuous active features using the SSR module. By converting meal and insulin events into continuous values, we aim to capture the dynamic relationships between these factors and blood glucose levels to improve the accuracy of our prediction model.

\subsection{Feature Decomposition Module (VMD)}
Variational Mode Decomposition (VMD) is an adaptive, non-wavelet multi-resolution decomposition technique that decomposes a complex signal into a set of band-limited intrinsic mode functions. VMD is an improved version of Empirical Mode Decomposition (EMD) and does not decompose the time series data with fixed functions. However, unlike EMD, VMD is a non-recursive algorithm that concurrently extracts multiple modes, with the number of modes ($m$) determined by the user based on the complexity of the problem~\cite{Liu2022ARO}. These $m$ modes are identified by signal peaks in the frequency domain. Each mode is centered around a specific pulsation frequency ($\omega_m$), which is essential for accurate analysis~\cite{6655981}. 

For a time series $X(t)$ where $X$ is a vector of size $T$, VMD decomposes the time series into $m$ modes, resulting in a new matrix $X_{new}$ of size $m \times T$, where $m$ presents the number of modes. we extend the VMD decomposition technique proposed in~\cite{6655981} to decompose the blood glucose time-series data to decrease the complexity of this signal. The VMD formulation for a single time series $X(t)$ can be expressed as:
\begin{equation}
\min_{\{u_k\},\{\omega_k\}} \left\{ \sum_{k=1}^m \left\| \partial_t \left[ \left(\delta(t) + \frac{j}{\pi t}\right) * u_k(t) \right] e^{-j\omega_k t} \right\|_2^2 \right\}
\end{equation}
subject to $\sum_{k=1}^m u_k = X(t)$, where $u_k(t)$ represents the decomposed modes, $\omega_k$ are the center frequencies for each mode, $\delta(t)$ is the Dirac distribution, and $*$ denotes convolution.

To solve this optimization problem, it is converted into an unconstrained Lagrangian form:

\small
\begin{equation}
\begin{aligned}
\mathscr{L} & (\{u_k\},\{\omega_k\}, \lambda):=\alpha \sum_{k}\left\|\partial_{t}\left[\left(\delta(t)+\frac{j}{\pi t}\right) * u_k(t)\right] e^{-j \omega_k t}\right\|_{2}^{2}\\
&+\left\|X_1(t)-\sum_{k} u_k(t)\right\|_{2}^{2} 
 +\left\langle\lambda(t), X_1(t)-\sum_{k} u_k\right\rangle
\end{aligned}
\end{equation}
\normalsize
\noindent where $\mathscr{L}$ is the Lagrangian function, $\lambda$ represents the Lagrange multipliers, and $\alpha$ is the balancing parameter for the data-fidelity constraint.

The Alternate Direction Method of Multipliers (ADMM) is employed to solve this problem iteratively. The updates for each mode, center frequency, and Lagrange multiplier are performed in the frequency domain as follows~\cite{6655981}:

\begin{equation}
\begin{aligned}
& \hat{\omega}_k^{i+1}=\frac{\int_{0}^{\alpha} \omega|\hat{u}_k(\omega)|^2 d \omega}{\int_{0}^{\alpha}|\hat{u}_k(\omega)|^2 d \omega} \\
& \hat{\lambda}^{i+1}(\omega)=\hat{\lambda}^n(\omega)+\tau(\hat{X_1}(\omega)-\sum_{k} \hat{u}_k^{n+1}(\omega)) \\
& u_k^{i+1}(t)=\operatorname{Re} \mathscr{F}^{-1}\left(\frac{\hat{X_1}(\omega)-\sum_{i \neq k} \hat{u}_i+\frac{\hat{\lambda}(\omega)}{2}}{1+2 \alpha(\omega-\omega_k)^2}\right)
\end{aligned}
\end{equation}

Detailed explanations of VMD are presented in~\cite{6655981}. Note that, in our proposed method, we apply the VMD to the training dataset and testing dataset separately to avoid information leakage from the test dataset to the training dataset. After applying VMD on data, the decomposed blood glucose signals are divided into two groups. The first group has less fluctuation and includes the trend of the signal, while the second group has higher fluctuation. The first group's modes are summed up to generate a new signal with a trend and low fluctuation of blood glucose values called low-frequency BGL ($x_{g}^{low}$). The second group's modes are also summed up to generate a new signal with high fluctuation of blood glucose features called high-frequency BGL ($x_{g}^{high}$). 
Both $x_{g}^{low}$ and $x_{g}^{high}$ are then combined with effective carbs intake ($X_c$) and insulin dosage ($X_{ins}$), which are computed in our SSR module (Section~\ref{Mathematical}), to create two new datasets. These two datasets are called low-frequency features dataset ($LFFD$) and high-frequency features dataset ($HFFD$), respectively.


\subsection{Forecasting Module}

\textcolor{black}{Blood glucose time series exhibit both slow-varying physiological trends and rapid transient fluctuations driven by meals, insulin, and behavioral factors. These components involve distinct temporal dependencies. The LSTM module is utilized to model the smooth, long-term dynamics associated with the low-frequency component, while the Transformer module captures the abrupt short-term variations inherent in the high-frequency component. Therefore, using these two lightweight models in parallel enables GlucoNet to exploit the complementary strengths of recurrent and attention-based architectures, leading to more robust multi-horizon forecasting.} Consequently, both $LFFD$ and $HFFD$ are then fed to the forecasting models separately to predict the blood glucose levels.
The forecasting employs a one-step model for various ranges of prediction horizons (PHs) such as Short-range ($5 minutes$), medium-range ($30 minutes$), and long-range ($60 minutes$). A CNN-LSTM model is used for forecasting $LFFD$. This model consists of a one-dimensional convolutional neural network (1D-CNN) followed by LSTM layers, where the 1D-CNN comprises two convolutional layers. The reason behind selecting this model is that the $LFFD$ has lower fluctuation compared to $HFFD$. Therefore, forecasting these data is possible using a model such as CNN-LSTM. This model is called the low-frequency forecasting model.
The low-frequency forecasting model architecture consists of a series of layers designed to process and analyze time-series data for blood glucose prediction. The input to the model is a windowed sample of past data with a length of $180$ minutes, which is determined to be the optimal input length for all prediction horizons. The input data first passes through a stack of one-dimensional convolutional and pooling layers. This initial stage is designed to extract relevant features from the time-series data. We used two groups of two back-to-back convolutional layers to extract important features based on their relevance to the prediction task.  Following the convolutional layers, the extracted features are flattened into a single vector. This vector is then fed into an LSTM (Long Short-Term Memory) block, which can \textcolor{black}{be} made up of different LSTM layers with different numbers of memory cells or units. These settings offer multiple implementation configurations for the low-frequency forecasting model. Therefore, considering a set of numbers of memory cells can provide flexibility in adapting GlucoNet to various accuracy requirements and computational constraints. In our low-frequency forecasting model of GlucoNet, each configuration of the model has $k$ LSTM layers. Each layer is defined using an input number of memory cells and an output number of memory cells configuration vector, $L_{vec}=[(L_{1,1},L_{1,2}),(L_{2,1},L_{2,2}),...,(L_{k,1}, L_{k,2})]$. Here, $(L_{i,1},L_{i,2})$ represents the input number of memory cells and the output number of memory cells in the $i^{th}$ LSTM layer. Hence, the generic low-frequency forecasting model of GlucoNet representation, GlucoNet $\{[(L_{1,1},L_{1,2}),(L_{2,1},L_{2,2}),...,(L_{k,1},L_{k,2})]\}$, defines some possible low-frequency forecasting model of GlucoNet configuration. Note that we considered two layers for our convolutional layers and two fully connected layers in the low-frequency model to decrease the size of design space and make it feasible to explore our model's configuration. For instance, one of our low-frequency forecasting models of  GlucoNet configurations has an LSTM layer with a $128$ input number of memory cells and $64$ output number of memory cells represented by GlucoNet $\{[(128,64)]\}$. Then, the LSTM block output is fed into two fully connected layers. The number of neurons in these layers varies depending on the prediction horizon, e.g., for $5-minute PH: [64\times32, 32\times1], 30-minute PH: [64\times32, 32\times6],$ and $60-minute PH: [64\times32, 32\times12]$ neurons. The low-frequency model architecture is able to progressively refine the extracted features and forecast BGL accurately for different time horizons.

\begin{figure}[h]
    \centering
    \includegraphics[width=0.7\linewidth]{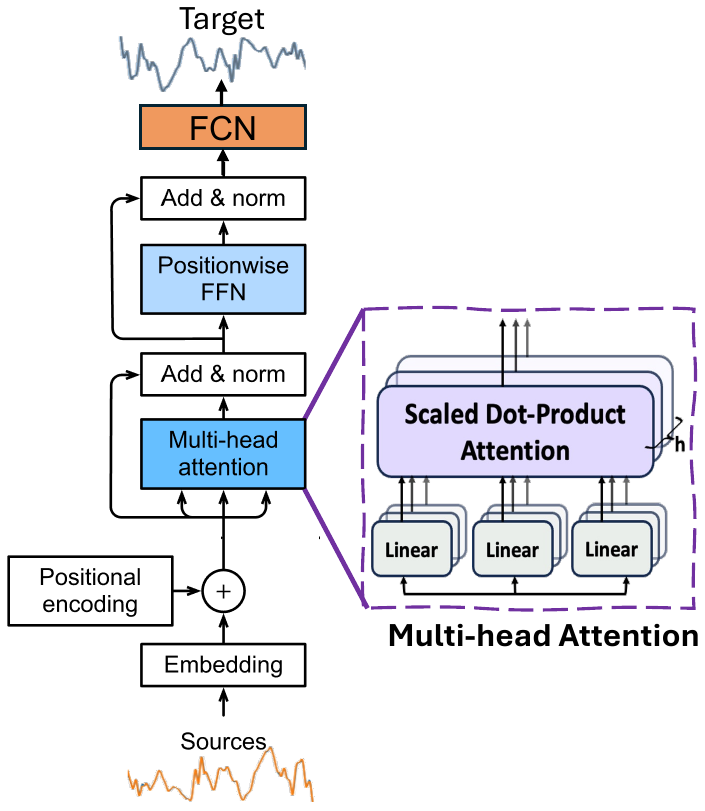}
    \caption{Overall Transformer Encoder model architecture for the teacher model \{$64$ input dimensions, $4$ attention heads, $128$ feed-forward units\} and student model \{$32$ input dimensions, $2$ attention heads, $64$ feed-forward units\}. }
    \label{fig:transformer}
\end{figure}

Due to high fluctuations in $HFFD$, we need to propose a complex model to forecast high-frequency blood glucose levels. \textcolor{black}{For forecasting high-frequency glucose fluctuations, GlucoNet employs the encoder block of a Transformer architecture. Only the encoder is used, as the supervision setting of the forecasting task does not require an autoregressive decoder. The encoder comprises multi-head self-attention, position-wise feed-forward sublayers, normalization layers, and residual connections, enabling the model to capture both short-term and irregular variability. In contrast, the low-frequency component is modeled using a compact CNN-LSTM network, which specializes in extracting long-term physiological patterns.}
The Transformer, developed by Google, introduced a novel approach to sequence modeling by replacing the traditional RNN structure with self-attention~\cite{vaswani2017attention}. By considering self-attention, Transformers addresses the potential issues in RNN regarding long-sequence dependencies and inefficiencies in backpropagation through time. In order to maintain the order of elements in a sequence, the Transformer uses positional encoding, especially for time-dependent tasks. Therefore, this adaptation makes it suitable for time series applications such as blood glucose monitoring, where tracking important temporal features can enhance the detection of irregular glucose levels ~\cite{lee2023glucose}.
The Transformer is made up of the encoder and the decoder. The encoder transforms input data into a set of contextual representations using self-attention followed by a multi-layered feed-forward network to capture complex relationships across the sequence. Afterward, the output from the encoder is fed to the decoder, which generates an output sequence. This process involves self-attention to identify relevant aspects of the generated output.
Taking advantage of these capabilities, in this study, we used the Transformer to forecast high-frequency fluctuations in blood glucose, which is hard to predict due to the rapid and unpredictable changes in glucose levels. We modified this design for our forecasting model and framed it as a supervised learning task. Specifically, we eliminated the decoder and only utilized the encoder for data representation learning. The overall Transformr architecture used for this study is shown~\ref{fig:transformer}. However, the complexity of transformers makes these models \textcolor{black}{are} unsuitable for real-time applications such as forecasting blood glucose.

\textcolor{black}{Blood glucose signals consist of slow-varying physiological trends and rapid high-frequency fluctuations driven by meals, insulin, and behavioral factors. These components exhibit different temporal structures, making a single forecasting model prone to learning conflicts, reduced stability, and suboptimal accuracy. By decomposing the CGM signal into low-frequency and high-frequency modes using VMD, each sub-signal becomes more stationary and predictable. This enables the use of two lightweight specialized models, including a compact LSTM for long-term trends and a small Transformer for rapid fluctuations. Moreover, the two models operate in parallel, preserving real-time feasibility with limited impact on inference latency. Since each model focuses on a narrower temporal behavior, their parameter sizes remain small, and the total parameter count of GlucoNet is substantially lower than existing architectures. Additionally, knowledge distillation is applied only to the high-frequency module to further compress the Transformer without sacrificing accuracy. This design yields an improved accuracy–efficiency trade-off while maintaining suitability for embedded AI deployment.}

Knowledge Distillation (KD) is a technique that facilitates the transfer of knowledge from one model, typically a larger teacher model, to another, usually a smaller student model. We used the benefit of using knowledge distillation approaches for training deep learning models, particularly in forecasting tasks. Knowledge distillation allows a student model to learn from a more complex teacher network. This process allows the student to learn from more informative sources, such as the predictive probabilities provided by the teacher. Hence, the student can achieve performance levels comparable to the teacher despite being a more compact model. In some cases, when the student and teacher have similar capacities, the student may even outperform the teacher~\cite{hinton2015distilling}. Therefore, we extended Knowledge Distillation (KD) to compact the large Transformer (Teacher model) to the small Transformer (Student model) to achieve more accurate forecasting of blood glucose levels for high-frequency features.  

For an input $\mathbf{x}$ and a $K$-dimensional target $\mathbf{y}$, a model generates a vector logits $\mathbf{z}(\mathbf{x}) = [z_1(\mathbf{x}), \ldots, z_K(\mathbf{x})]$, which is then converted into predicted probabilities $P(\mathbf{x}) = [p_1(\mathbf{x}), \ldots, p_K(\mathbf{x})]$ using a softmax function. Hinton et al.~\cite{hinton2015distilling} proposed temperature scaling to soften these probabilities for improved distillation using Eq.~\ref{eq:softmax}.

\begin{equation}
\widetilde{p}_i(\mathbf{x}; \tau) = \frac{\exp(z_i(\mathbf{x}) / \tau)}{\sum_j \exp(z_j(\mathbf{x}) / \tau)}
\label{eq:softmax}
\end{equation}

\noindent where $\tau$ is the temperature parameter. By adjusting the softmax output $P^T(\mathbf{x})$ of the teacher and $P^S(\mathbf{x})$ of the student, the student is trained using a combination of teacher probabilities and ground truth labels. Eq.~\ref{eq:lossstudentfunction} presents the loss function ($\mathcal{L}_{KD}$) used for training the student model.

\begin{equation}
\begin{aligned}
\mathcal{L}_{KD}(\mathbf{x}, \mathbf{y}) = & (1-\alpha) H(\mathbf{y}, P^S(\mathbf{x})) + \\
& \alpha \tau^2 H(\widetilde{P^T}(\mathbf{x}; \tau), \widetilde{P^S}(\mathbf{x}; \tau))
\end{aligned}
\label{eq:lossstudentfunction}
\end{equation}

\noindent where $H$ denotes Mean Squared Error (MSE) loss and $\alpha$ is a hyperparameter. 

It is worth noting that, in some cases, the student model may outperform the teacher after knowledge distillation. This behavior has been observed in prior studies and is attributed to improved generalization. The student benefits from both reduced model capacity, which limits overfitting, and the additional supervision provided by softened output distributions, which encode richer relational information. As a result, the distilled student can learn more robust representations and achieve better predictive performance than the teacher, particularly in settings with noisy or limited data, such as CGM-based glucose forecasting.

Knowledge distillation techniques address the issues related to using a complex model to achieve an acceptable performance in forecasting blood glucose model. In this regard, we proposed a large teacher model as a Transformer and a small student model to forecast blood glucose levels. The overall architecture of the teacher and student model is illustrated in Fig.~\ref{fig:transformer}. 

We conducted a series of experiments to optimize the Transformer model architecture.
For the teacher model, the best-performing configuration consisted of a single encoder layer, $64$ input dimensions, $4$ attention heads, and $128$ feed-forward units. 
Moreover, for the student model, the optimal configuration differed slightly. It utilized a single encoder layer, $32$ input dimensions, $2$ attention heads, and $64$ feed-forward units. These setups demonstrated superior performance compared to other configurations. Also, the high-frequency model of the proposed GlucoNet offers multiple implementation configurations. It can be deployed with or without knowledge distillation (KD) and implemented with the large Transformer (teacher Transformer) $(LT)$ or the small transformer (student Transformer) $(ST)$. These options provide flexibility in adapting GlucoNet to various accuracy requirements and computational constraints.
Thus, the generic GlucoNet representation, GlucoNet $(+,)$ (KD,) $(ST, LT)$ $\{[(L_{1,1},L_{1,2}),(L_{2,1},L_{2,2}),...,(L_{k,1},L_{k,2})]\}$,  defines some possible GlucoNet configuration. For example, GlucoNet $+$ KD $(ST)$ $\{[(32,16)]\}$  refers to a model that is made up of one LSTM layer with 32 input numbers of memory cells and 64 output numbers of memory cells, and implemented by KD with small Transformer as a student model and GlucoNet $(LT)$ $\{[(32,16)]\}$  refers to a model that is made up of one LSTM layer with 32 input numbers of memory cells and 64 output numbers of memory cells, followed by large Transformer (without KD).  

Eventually, the outputs of the CNN-LSTM model and the Transformer are added to construct the forecasted blood glucose levels.

In summary, accurate forecasting of blood glucose levels for different ranges of PHs by GlucoNet is presented by Algorithm~\ref{alg:vmd_kd}. First, CGM time series data are decomposed by using the VMD technique to low-frequency and high-frequency features (line $7$ of the Algorithm~\ref{alg:vmd_kd}), and then the event-based features, such as carbohydrate intake and insulin dosage, are transformed into continuous time series data (line $9$ of the Algorithm~\ref{alg:vmd_kd}). Then, the continuous time series data are combined by low frequency and high-frequency features and generate two sets of features. The LSTM model is trained by using $LFFD$ (line $10$ of the Algorithm~\ref{alg:vmd_kd}), and the transformer-based is trained by $HFFD$ (lines $11$ and $12$ of the Algorithm~\ref{alg:vmd_kd}).  The $LFFD$ are fed to the LSTM model to predict the low frequency of blood glucose levels (line $15$ of the Algorithm~\ref{alg:vmd_kd}), and $HFFD$ are fed to the knowledge distillation transformer-based model to forecast the high-frequency blood glucose levels (line $16$ of the Algorithm~\ref{alg:vmd_kd}). Finally, the predicted blood glucose levels for a PH are obtained by combining the low-frequency and high-frequency blood glucose levels (line $17$ of the Algorithm~\ref{alg:vmd_kd}).
\begin{algorithm}[h]
    \small
   \caption{\small 
   GlucoNet for BGL forecasting.
   Here, $X_g, X_{\text{ins}}, X_c \in \mathbb{R}^{36 \times 1}$, $TF()$ is a mathematical model for feature transformation, $T$: teacher model, $S$: student model.}
   \label{alg:vmd_kd}
\begin{algorithmic}[1]
   \STATE {\bfseries Input:} Training data $X_{\text{train}}, Y_{\text{train}}$; Test data $X_{\text{test}}, Y_{\text{test}}$; Features $X = \{ X_g, X_{\text{ins}}, X_c \}$; Define GlucoNet configuration
   \STATE {\bfseries Output:} Error metrics (MSE, RMSE, $R^2$)
   \STATE {\bfseries Begin}
   \STATE Initialize LSTM parameters $\theta_{\text{LSTM}}$, Student model parameters $\alpha_S$, Teacher model parameters $\gamma_T$

   \STATE {\bfseries Training:}
   
   \FORALL{samples $X_i \in X_{\text{train}}$}
        \STATE $X_{L,i}, X_{H,i} \leftarrow \text{VMD}(X_{g,i})$ \quad \COMMENT{Decompose CGM signal}
        \STATE $X_{L,i}^N = \text{$norm$}(X_{L,i})$ \quad 
        \STATE $X_{\text{ins},c} = TF(X_{\text{ins},i}, X_{c,i})$ \quad \COMMENT{Transform sparse features to continuous}
        \STATE $\theta_{\text{LSTM}} \leftarrow \arg \min_{\theta} L(\theta; X_{\text{ins},c}, X_{L,i}^N, Y_{\text{train}})$ \quad \COMMENT{Train LSTM}
        \STATE $\gamma_{\text{T}} \leftarrow \arg \min_{\gamma} L(\gamma; X_{\text{ins},c}, X_{H,i}^N, Y_{\text{train}})$ \quad \COMMENT{Train Teacher}
        \STATE $\alpha_S \leftarrow \arg \min_{\alpha} L(\alpha;\gamma; X_{\text{ins},c}, X_{H,i}, Y_{\text{train}})$ \quad \COMMENT{Knowledge Distillation}
    \ENDFOR
    \STATE {\bfseries Inference:}
        \STATE $Y_1 \leftarrow \theta_{\text{LSTM}}(X_{\text{ins},c}^{\text{test}}, X_L^{\text{test}}, Y_{\text{test}})$ \quad \COMMENT{Prediction on low-frequency data}
        \STATE $Y_2 \leftarrow \alpha_S(X_{\text{ins},c}^{\text{test}}, X_H^{\text{test}}, Y_{\text{test}})$ \quad \COMMENT{Prediction on high-frequency data}
        \STATE $\hat{Y} = \text{$Denorm$}(Y_1) + Y_2$ \quad \COMMENT{Combine predictions}
    
    \STATE {\bfseries return} $\text{RMSE}, \text{MAE}, R^2\leftarrow \text{$Error$}(\hat{Y}, Y_{\text{test}})$
    \STATE {\bfseries End}
\end{algorithmic}
\end{algorithm}
\vspace{-0.4cm}
\color{black}
\section{Experimental Evaluation}
\label{sec:Experimental_Evaluation}
We evaluate the performance of GlucoNet and compare it with a broad range of state-of-the-art blood glucose forecasting models to demonstrate its effectiveness. Performance is assessed using standard error metrics (e.g., RMSE and MAE), as well as model complexity, measured by the total number of parameters and inference latency on an edge device.
The state-of-the-art blood glucose forecasting models used for comparison include an LSTM-based model proposed in \cite{martinsson2020blood}, a convolutional and recurrent neural network (CNN–RNN) proposed in \cite{freiburghaus2020deep}, the 1D CNN–LSTM GlySim model \cite{arefeen2023glysim}, a CNN–RNN model proposed in \cite{daniels2021multitask}, MTL-LSTM \cite{shuvo2023deep}, as well as recent state-of-the-art glucose forecasters, including TimesFM \cite{das2024decoder}, GLIMMER \cite{glimmer}, and GlyRAG \cite{glyRag}. Notably, GLIMMER employs a CNN–LSTM architecture trained with a dysglycemia-weighted loss function to emphasize clinically critical glucose ranges, while GlyRAG leverages large language models for context-aware, retrieval-augmented forecasting. Furthermore, we include recent transformer-based models for comparison, including Sequential-T~\cite{11250578}, Transformer used in~\cite{karagoz2025comparative}, Crossformer~\cite{zhang2023crossformer}, PatchTST~\cite{nie2022time}, iTransformer~\cite{liu2023itransformer},  TimeXer~\cite{wang2024timexer}, and Gluformer~\cite{sergazinov2023gluformer}. These models are evaluated on the OhioT1DM dataset using consistent input features (CGM, insulin, and carbohydrate intake), ensuring a fair and consistent comparison across models and prediction horizons with the proposed GlucoNet configuration.

GlucoNet is evaluated on two independent Type 1 Diabetes Mellitus (T1DM) datasets to assess both prediction accuracy and generalization across cohorts. Moreover, we analyze the accuracy–efficiency trade-off of the proposed architecture, conduct ablation studies to isolate the contributions of architectural components, and evaluate robustness across datasets, demographic groups, and clinically critical glycemic events. Details of the datasets used in this evaluation are provided below.

\subsection{Dataset}
\label{dataset}
\subsubsection{OhioT1DM}
The dataset that we used to demonstrate our proposed method is OhioT1DM (2018 and 2020 versions)~\cite{marling2020ohiot1dm}.
This is a clinical dataset with heterogeneous data, including diverse age, gender, and device variability, that reduces the risks of underfitting to deal with unseen data. Each version consists of comprehensive data on six individuals, which are a total of 12 participants (7 male, 5 female) with type 1 diabetes over an eight-week period. The cohort comprised two male and four female participants. Data collection was facilitated through Medtronic 530G insulin pumps and Medtronic Enlite CGM sensors, which were employed consistently throughout the study duration. The dataset encompasses various measurements such as Continuous Glucose Monitoring (CGM) readings at 5-minute intervals, insulin administration, including both bolus and basal doses, and self-reported data on meal times with estimated carbohydrate intake information. Moreover, other information includes exercise (E), sleep, work, stress, and illness. A summary of the dataset is provided in Table~\ref{tab:dataset}. Note that the missing values, especially for CGM values, are imputed with interpolation.

\subsubsection{AZT1D}
To assess generalization to contemporary automated insulin delivery (AID) use, we also analyze AZT1D~\cite{khamesian2025azt1d}, a new clinic-sourced dataset collected at Mayo Clinic (Scottsdale, AZ) between Dec 2023–Apr 2024. The cohort includes 25 adults (13 female, 12 male; age 27–80, mean 59), each contributing on average ~26 days of real-world data. AZT1D contains Dexcom G6 Pro CGM (5-min sampling), Tandem t:slim X2 pump logs (including granular bolus details: total dose, bolus type, correction components), carbohydrate intake, and device mode (regular/sleep/exercise). In total, the release comprises 320,488 CGM readings spanning ~26,707 hours. Compared with OhioT1DM, AZT1D offers richer therapy context under AID, enabling studies on morphology-aware forecasting, decision support, and patient-twin personalization. 

\begin{table}[t]
\centering
\caption{Summary of OhioT1DM Dataset}
\begin{tabular}{|c|c|c|c|c|c|}
\hline
\textbf{Year} & \textbf{Gender} & \textbf{PID} & \textbf{\# Training} & \textbf{\# Testing} & \textbf{Age} \\
& & & \textbf{Samples} & \textbf{Samples} & \textbf{(years)} \\
\hline
\multirow{6}{*}{2018} & \multirow{4}{*}{Female} & 591 & 10847 & 2760 & \multirow{8}{*}{40--60} \\
& & 588 & 12640 & 2791 & \\
& & 575 & 11866 & 2590 & \\
& & 559 & 10796 & 2514 & \\
\cline{2-5}
& \multirow{2}{*}{Male} & 570 & 10982 & 2745 & \\
& & 563 & 12124 & 2570 & \\
\cline{1-5}
\multirow{6}{*}{2020} & \multirow{5}{*}{Male} & 544 & 10623 & 2704 & \\
& & 584 & 12150 & 2653 & \\
\cline{6-6}
& & 569 & 10877 & 2731 & 60--80\\
\cline{6-6}
& & 540 & 11947 & 2884 & \multirow{3}{*}{20--40}\\
& & 552 & 9080 & 2352 &\\
\cline{2-5}
& Female & 567 & 10947 & 2377 &\\
\hline
\end{tabular}
\label{tab:dataset}
\end{table}

\subsection{Performance Metrics}
In this work, we used (1) Root Mean Square Error (RMSE)~\cite{arefeen2023glysim}, (2) Mean Absolute Error (MAE)~\cite{arefeen2023glysim}, and (3) $R^2$ Square~\cite{syafrudin2022personalized,khadem2023blood} as the error metrics for comparing the accuracy of the different forecasting model. The definitions of these error metrics are presented below.

\textbf{Root Mean Square Error (RMSE)} of forecasted values of blood glucose levels for specific Prediction Horizons (PHs) is defined as:
\begin{equation}
\text{RMSE} = \sqrt{\frac{1}{n} \sum_{i=1}^{n} (y_i - \hat{y_i})^2}
\end{equation}
Here, $y_i$ and $\hat{y_i}$ refer to the actual value and the predicted value, respectively.
$n$ refers to the number of test samples.

\textbf{Mean Absolute Error (MAE)} of forecasted values of blood glucose levels for specific Prediction Horizons (PHs) is defined as:
\begin{equation}
\text{MAE} = \frac{1}{n} \sum_{i=1}^{n} |y_i - \hat{y_i}|
\end{equation}

$\mathbf{R^2}$ of forecasted values of blood glucose levels for specific Prediction Horizons (PHs) is defined as: 
\begin{equation}
   R^2 = 1 - \frac{\sum_{i=1}^N (y_i - \hat{y_i})^2}{\sum_{i=1}^N (y_i - \overline{y_i})^2} 
\end{equation}
Here, $\overline{y_i}$ presents the mean of the actual BGL.

\subsection{Comparison GlucoNet with State-of-the-Art (SOTA)}

We compared our proposed GlucoNet model with SOTA models by implementing each model on the OhioT1DM and AZ1TD datasets and comparing their performance using error metrics. 
\subsubsection{Experimental Setup}
The data from each dataset is split into training and test sets for each participant. The models are trained on the training dataset. Prediction metrics are computed using the test dataset for each participant. The model's input consists of a three-hour ($180$ minutes) sliding window of historical data, providing sufficient information for making good predictions for the next prediction horizon (PH). Various PHs are considered for comparison, which are $5$ minutes, $30$ minutes, and $60$ minutes.
The model architecture and hyperparameters are selected based on prior work in blood glucose forecasting and time-series modeling, followed by a systematic hyperparameter optimization procedure using a held-out validation set, with additional validation through ablation studies. Key parameters, including the input window length, number of VMD modes, LSTM hidden units, and Transformer attention heads, are chosen to improve RMSE and MAE while maintaining generalization and computational efficiency for edge deployment.

The LSTM model was trained for 300 epochs,
repeated over 5 separate runs to ensure consistency and reliability of results. Moreover, our teacher model is trained for 500 epochs; then, the student model is trained by the knowledge distillation loss method for 500 epochs as well. 

The training process began with an initial learning rate of $0.001$. The adaptive moment estimation (Adam) optimizer is chosen to minimize the linear combination of the MSE loss function in both LSTM and teacher models. The student model used the Adam optimizer to minimize the knowledge distillation loss function. Adam optimizer is selected due to its suitability for learning in non-stationary data, such as blood glucose.
A minibatch training approach was adopted, using a batch size of 64.


\subsubsection{Comparative Evaluation on the OhioT1DM Dataset}

We compared GlucoNet with SOTA models. Table~\ref{tab:performance-compare} presents a comparison of GlucoNet with other SOTA models in terms of RMSE, MAE, and $R^2$ Square for PHs of 30 minutes and 60 minutes. The GlucoNet configuration used for this comparison is GlucoNet $+$ KD $(ST)$ $\{[(128,64)]\}$. All error metrics reported in Table~\ref{tab:performance-compare} are averaged for fair comparison. The results show that GlucoNet consistently outperforms existing methods, highlighting the effectiveness of the proposed forecasting framework.

The SOTA models that we compared in Table~\ref{tab:performance-compare} included deep learning models such as long short-term memory (LSTM), 1D CNN–LSTM (GlySim), deep residual networks, recurrent self-attention models, convolutional recurrent neural networks (CNN–RNN), Transformer-based model and large language model–based forecasters. For $PH=30$ minutes, GlucoNet with KD achieved the lowest RMSE of $10.03 mg/dL$ and MAE of $6.95 mg/dL$. This outperformed the strongest transformer-based model, PatchTST, which had an RMSE of $15.61 mg/dL$ and an MAE of $9.23 mg/dL$. Therefore, this corresponds to about $36\%$ and $33\%$ improvement in RMSE and MAE, respectively.

At a prediction horizon of $PH=60$ minutes, GlucoNet+KD achieved an RMSE of $16.13 mg/dL$ and MAE of $10.67 mg/dL$, improving upon GlySim, which reports an RMSE of $24.2 mg/dL$ and MAE of $16.5 mg/dL$. This corresponds to approximately $33\%$ in RMSE and $36\%$ reductions in RMSE and MAE, respectively, compared with SOTA models.
Moreover, GlucoNet consistently outperforms recent foundation and LLM-based approaches, including TimesFM ($21.71/13.05 mg/dL$), GLIMMER ($23.97/15.83 mg/dL$), and GlyRAG ($20.22/12.33 mg/dL$), corresponding to $20–33\%$ lower RMSE and $13–33\%$ lower MAE. These results highlight the effectiveness of the proposed GlucoNet in capturing complex blood glucose dynamics and maintaining strong forecasting performance over longer prediction horizons.

\begin{table*}[t]
\caption{\small Comparison of a configuration of GlucoNet with KD with state-of-the-art Forecasting Blood Glucose models. The best values for error metrics are colored in red.}
    \centering
    \small
    \begin{tabular}{cc|c|*{2}{c}|*{2}{c}}
    \toprule
        \multicolumn{3}{c|}{\textbf{Prediction Horizon (PH)}} & \multicolumn{2}{c|}{\textbf{30 min}} & \multicolumn{2}{c}{\textbf{60 min}}\\
        \hline
         Dataset & Study & \textbf{Input Features} & RMSE & MAE & RMSE & MAE \\
         \midrule
         \multirow{4}{*}{\centering 2018} & LSTM~\cite{martinsson2020blood} & BGL & 18.86 & - & 31.40 & - \\
         & GluNet~\cite{li2019glunet} & BGL, Insulin, Carbs & 19.28 & - & 31.82 & -\\
         & MTL-LSTM~\cite{shuvo2023deep} & BGL, Insulin, Carbs & 15.73 & 10.43 & 30.01 & 21.36\\
         \cline{2-7}
         & \makecell{GlucoNet\small$+$KD, $\{(128,64)\}$} & BGL, Insulin, Carbs & \textcolor{red}{9.07} & \textcolor{red}{6.52} & \textcolor{red}{14.37} & \textcolor{red}{9.96} \\
        
         \midrule
         \multirow{5}{*}{\centering 2020} & CNN-RNN~\cite{freiburghaus2020deep} & BGL, Insulin, Carbs & 17.54 & 11.22 & 33.67 & 23.25\\
         & Deep Residual~\cite{rubin2020deep} & BGL, FG, BI, ToD, Carbs & 18.22 & 12.83 & 31.66 & 23.60\\
         & Knowledge Distillation~\cite{hameed2020investigating} & BGL, Insulin, Carbs & 19.21 & 13.08 & 31.77 & 23.09\\
         & MTL-LSTM~\cite{shuvo2023deep} & BGL, Insulin, Carbs & 16.39 & 10.86 & 31.78 & 22.77\\
         \cline{2-7}
         & \makecell{GlucoNet\small$+$KD, $\{(128,64)\}$} & BGL, Insulin, Carbs & \textcolor{red}{10.99} & \textcolor{red}{7.38} & \textcolor{red}{17.89} & \textcolor{red}{11.39} \\
        
        \midrule
        \multirow{15}{*}{\centering Combined two dataset} & Recurrent Self-Attention~\cite{cui2021personalised} & BGL, Insulin, Carbs & 17.82 & - & 28.54 & -\\
        & CNN-RNN~\cite{daniels2021multitask} & BGL, Insulin, Carbs, Exercise & 18.8 & 13.2 & 31.8 & 23.4\\
        & MTL-LSTM~\cite{shuvo2023deep} & BGL, Insulin, Carbs & 16.06 & 10.64 & 30.89 & 22.07\\
        & GlySim~\cite{arefeen2023glysim} & BGL, Insulin, Carbs & 17.5 & 12.3 & 24.2 & 16.5 \\
        & GLIMMER~\cite{glimmer} & BGL & - & - & 23.97 & 15.83\\
        & TimesFM~\cite{das2024decoder} & BGL & 11.46 & 6.69 & 21.71 & 13.05\\
        & GlyRAG~\cite{glyRag} & BGL, Context & 10.61 & 6.19 & 20.22 & 12.33\\
        & Transformer~\cite{karagoz2025comparative} & BGL, Insulin, Carbs & 27.71 & 22.24 & 31.19 & 24.55\\
        & Crossformer~\cite{zhang2023crossformer} & BGL, Insulin, Carbs & 16.12 & 10.11 & 24.73 & 16.30\\
        & PatchTST~\cite{nie2022time} & BGL, Insulin, Carbs & 15.61 & 9.23 & 24.89 & 15.98\\
        & iTransformer~\cite{liu2023itransformer} & BGL, Insulin, Carbs & 16.54 & 9.69 & 26.41 & 16.43\\
        & TimeXer~\cite{wang2024timexer} & BGL, Insulin, Carbs & 16.98 & 10.48 & 26.64 & 17.11\\
        & Gluformer~\cite{sergazinov2023gluformer} & BGL & 17.27 & 10.86 & 27.14 & 16.74\\
        & Sequential-T~\cite{11250578} & BGL, Insulin, Carbs & 16 & 9.45 & 28.99 & 16.83\\
        \cline{2-7}
        & \makecell{GlucoNet\small$+$KD, $\{(128,64)\}$} & BGL, Insulin, Carbs & \textcolor{red}{10.03} & \textcolor{red}{6.95} & \textcolor{red}{16.13} & \textcolor{red}{10.67} \\
        \midrule
    \end{tabular}
    \label{tab:performance-compare}
\end{table*}

\begin{figure*}[t]
\centering
    \subfloat[Predict low-frequency features]{
    \includegraphics[width=0.33\textwidth]{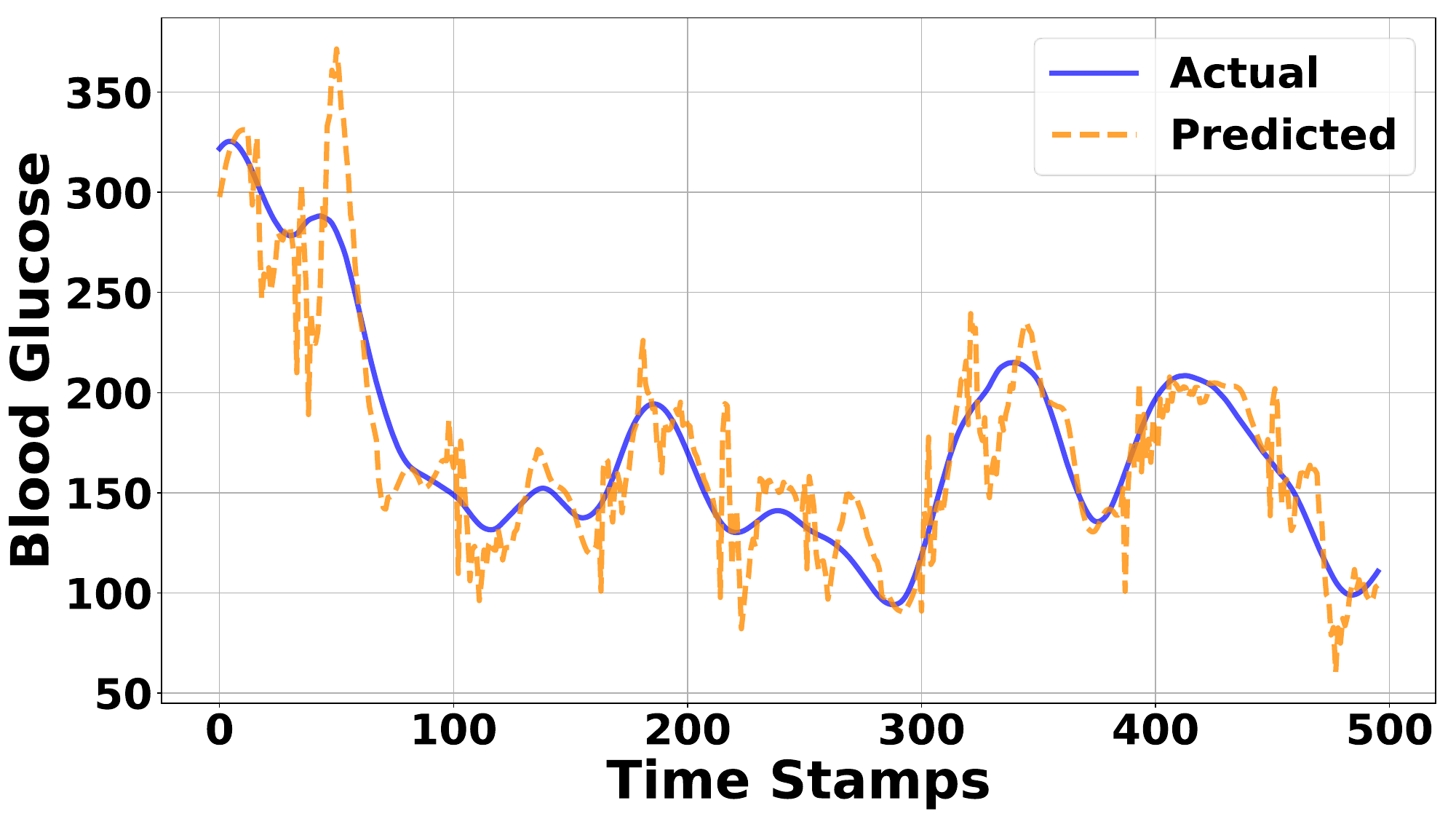}
    \label{predict_low}
    }
    \subfloat[Predict high-frequency features]{
    \includegraphics[width=0.33\textwidth]{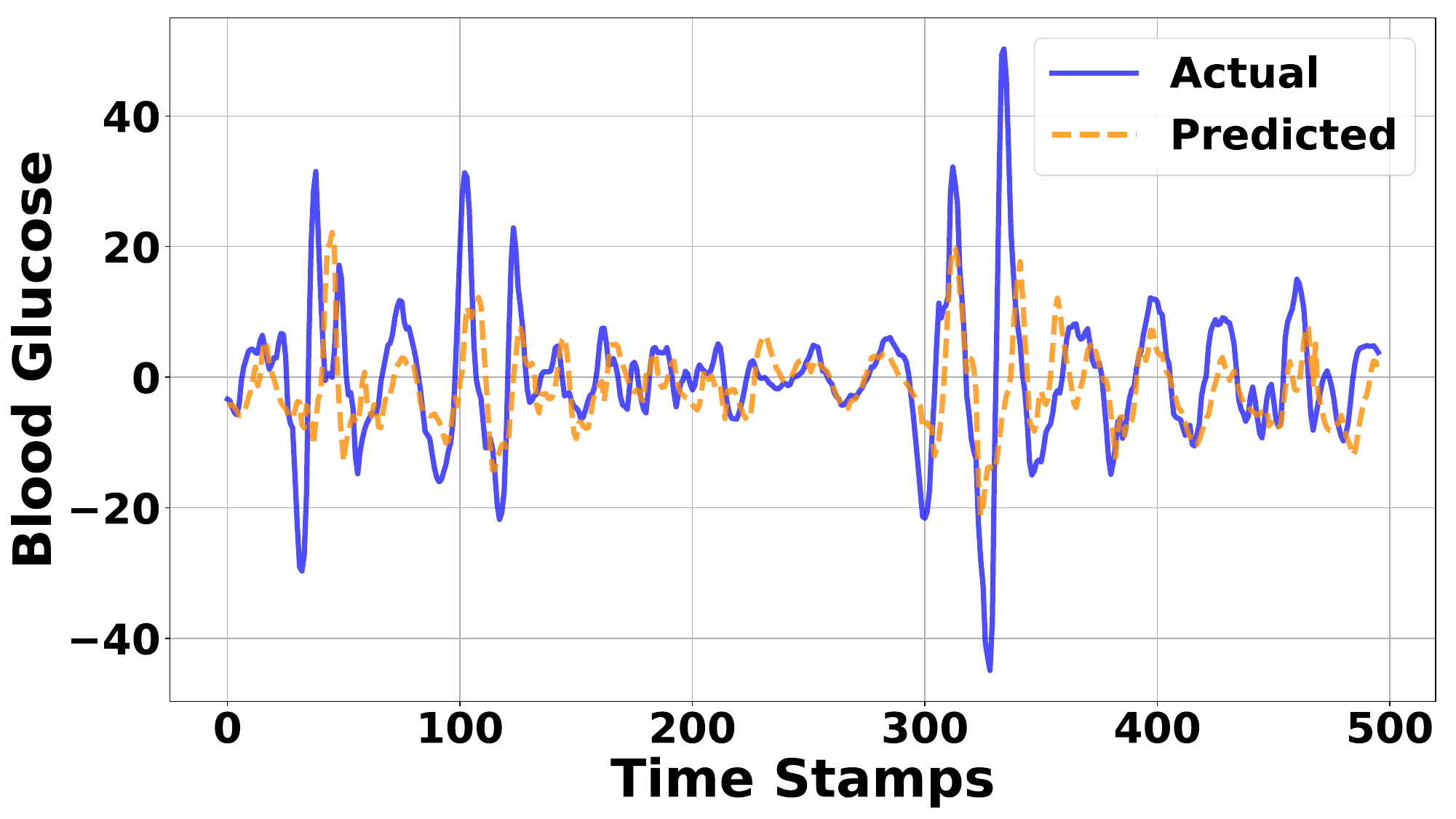}
    \label{predict_high}
    }
    \subfloat[Predict blood glucose level]{
    \includegraphics[width=0.33\textwidth]{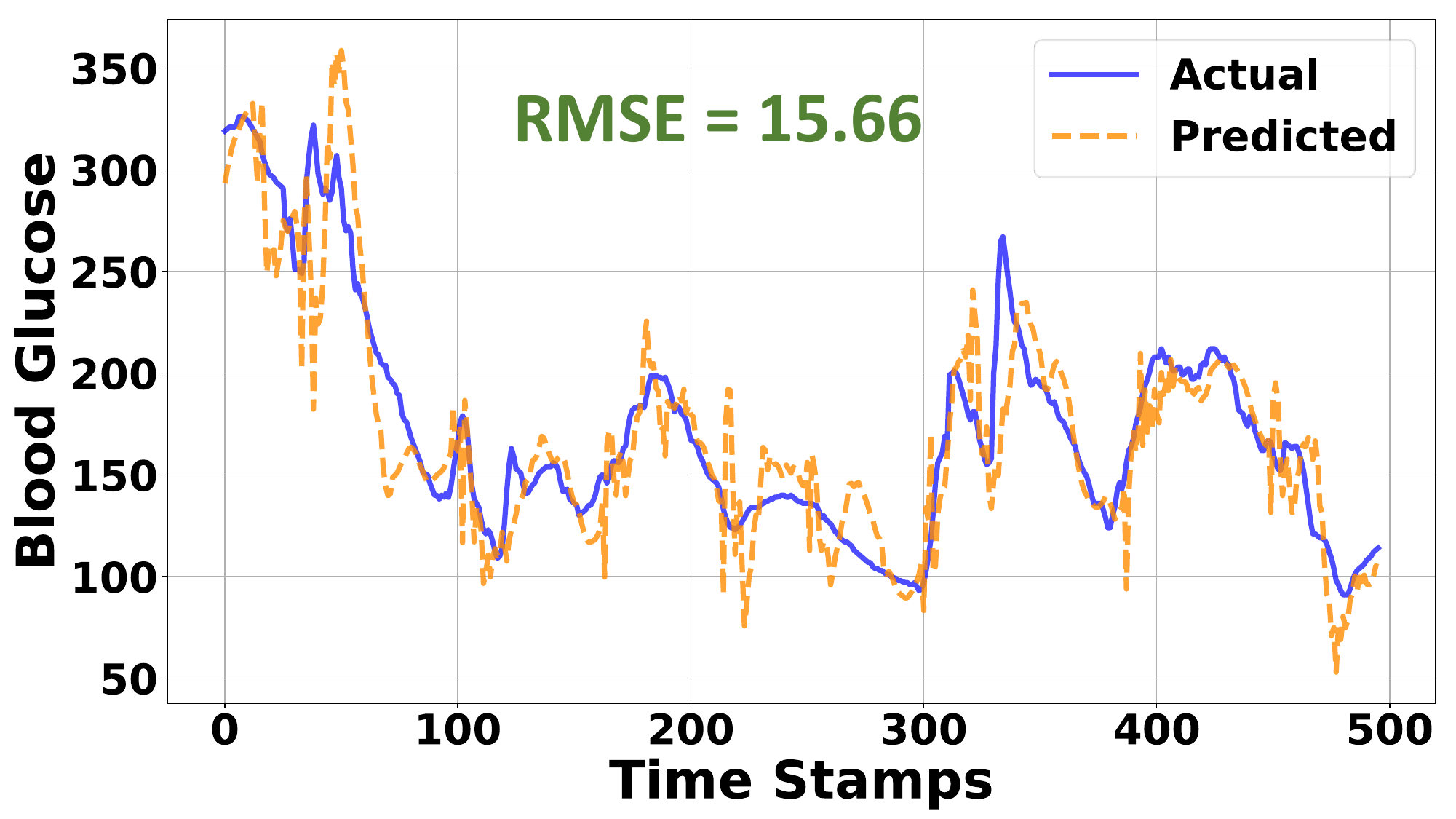}
    \label{predicted_tot}
    }
     \caption{Example of blood glucose forecasting with a prediction horizon of $PH=60$ minutes for a representative participant from the Ohio test dataset. (a) Low-frequency features prediction, (b) high-frequency features prediction, and (c) reconstructed blood glucose trajectory. The predicted and ground truth (Actual) signals are shown for visual comparison, and the RMSE for this subject is 15.66.}
     \label{fig:forecast}
\end{figure*}

An illustrative example of the blood glucose forecasting performance is presented in Fig.~\ref{fig:forecast}, which demonstrates GlucoNet’s predictions at a prediction horizon of $PH=60$ minutes using low-frequency features, high-frequency features, and their combined reconstruction for a representative participant from the Ohio dataset. As shown in Fig.~\ref{fig:forecast}, the predicted glucose trajectories closely follow the ground truth, indicating that GlucoNet maintains accurate forecasting performance even at a 60-minute horizon.


 Furthermore, the computational efficiency of the GlucoNet model compared to other models will be discussed in the following section.

\begin{figure}[h]
    \centering
    \includegraphics[width=0.95\linewidth]{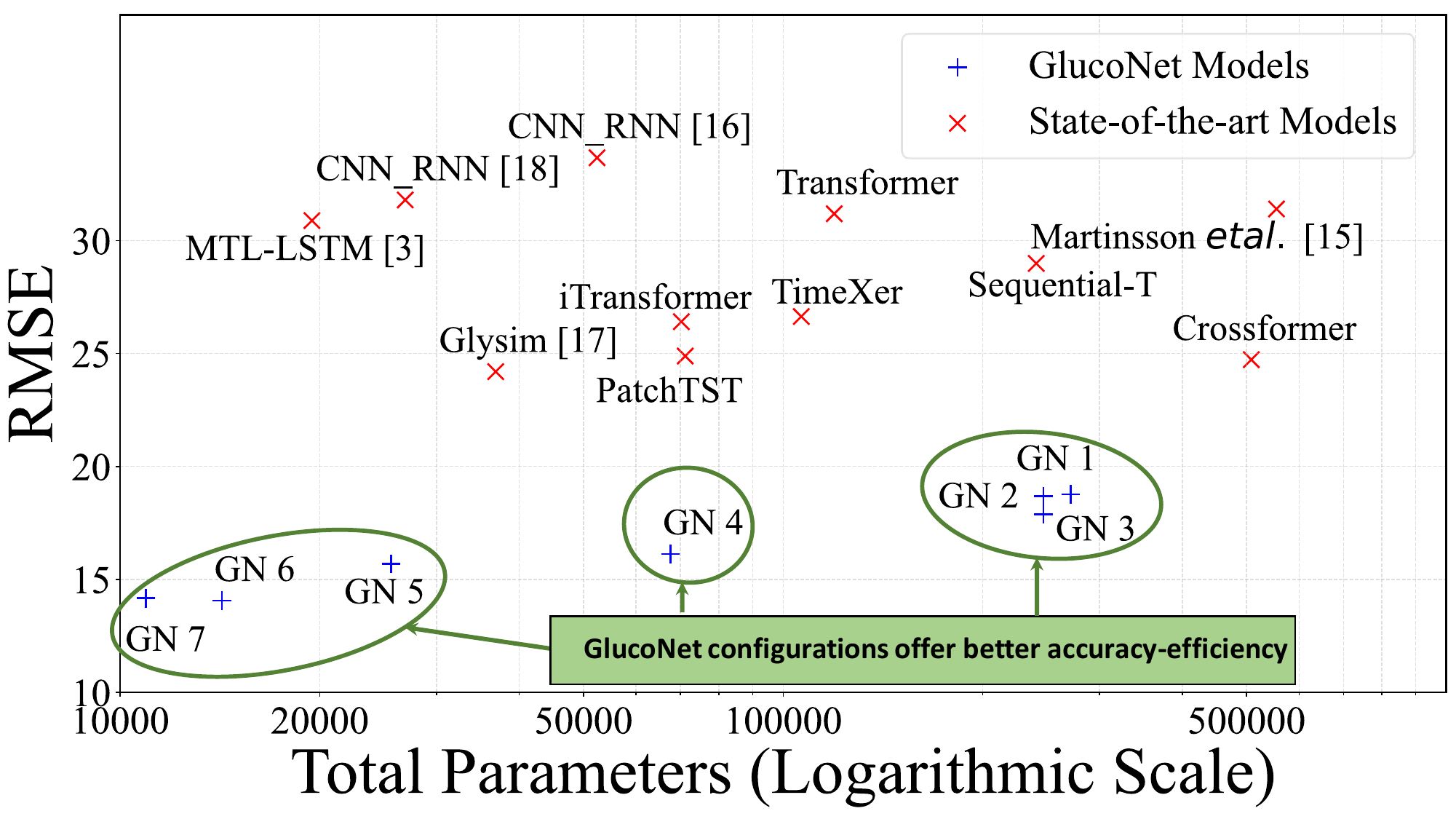}
    \caption{\small Accuracy-efficiency trade-offs for different configurations of GlucoNet compared to state-of-the-art works. GN 1: GlucoNet\small$+$KD \textit{(ST)}, $[(128,128),(128,64)]$, GN 2: GlucoNet\small \textit{(ST)}, $[(128,128),(128,64)]$, GN 3: GlucoNet \textit{(LT)}, $[(128,128),(128,64)]$, GN 4: GlucoNet\small$+$KD \textit{(ST)}, $[(128,64)]$, GN 5: GlucoNet\small$+$KD \textit{(ST)}, $[(64,32)]$, GN 6: GlucoNet\small$+$KD \textit{(ST)}, $[(32,16)]$, GN 7: GlucoNet\small$+$KD \textit{(ST)}, $[(16,8)]$}
    \label{fig:acc-eff trade}
\end{figure}

\subsubsection{Comparative Evaluation on the AZ1TD Dataset}
To evaluate generalization of GlucoNet, we extend our experimental evaluation to the AZT1D dataset.
GlucoNet is trained and evaluated on AZT1D using the same modeling pipeline and hyperparameter configurations adopted for OhioT1DM. Performance is benchmarked against two established baselines includes 1. a standard multimodal CNN–LSTM model and 2. GLIMMER~\cite{glimmer} model.
Table~\ref{tab:azt1d-comparison} summarizes the forecasting performance across 30-minute and 60-minute prediction horizons. The GlucoNet configuration with knowledge distillation (GlucoNet $+$ KD $(ST)$ $\{[(128,64)]\}$) consistently achieves the lowest prediction error across all horizons. At the 60-minute horizon, GlucoNet attains an RMSE of $14.12 mg/dL$ and an MAE of $9.70 mg/dL$, substantially improving upon GLIMMER ($22.48/15.58 mg/dL$) and CNN–LSTM ($29.55/21.61 mg/dL$).

\begin{table}[t]
\caption{\small Comparative evaluation on the AZT1D dataset across 30-minute and 60-minute prediction horizons.he best values
for error metrics are colored in red.}
\label{tab:azt1d-comparison}
\centering
\resizebox{\columnwidth}{!}{
\begin{tabular}{l|ccc|cc}
\toprule
\multirow{2}{*}{\textbf{Model}} &
\multicolumn{3}{c|}{\textbf{60 min}} &
\multicolumn{2}{c}{\textbf{30 min}} \\
\cline{2-6}
 & RMSE & MAE & $R^2$ & RMSE & MAE \\
\midrule
CNN--LSTM & 29.55 & 21.61 & 0.80 & 17.21 & 10.92 \\
GLIMMER & 22.48 & 15.58 & 0.81 & 14.46 & 9.39 \\
LSTM + Small Transformer & 14.20 & 9.84 & 0.86 & 9.36 & 6.69 \\
LSTM + Large Transformer & 14.81 & 10.43 & 0.83 & 9.98 & 7.31 \\
GlucoNet + KD (ST) \{[(32,16)]\} & \textcolor{red}{14.12} & \textcolor{red}{9.70} & \textcolor{red}{0.87} & \textcolor{red}{9.27} & \textcolor{red}{6.54} \\
\bottomrule
\end{tabular}
}
\end{table}

These results demonstrate that the performance gains achieved by GlucoNet are not dataset-specific and generalize effectively to a larger and more diverse patient cohort. The consistent accuracy improvements across prediction horizons validate the robustness of the proposed architecture and its applicability beyond benchmark datasets.

\subsection{Accuracy-Efficiency Trade-offs}
We have also compared our GlucoNet with other State-of-the-Art forecasting models in terms of accuracy-efficiency. We computed accuracy metrics, e.g., RMSE and MAE. Furthermore, the efficiency metric is computed by the total number of model parameters. Fig.~\ref{fig:acc-eff trade} shows the accuracy-efficiency trade-offs of GlucoNet compared with state-of-the-art models.
We can note that GlucoNet configurations offer the best accuracy-efficiency trade-off compared to the other models. Our GlucoNet with KD achieves high accuracy with only about $10,900$ total parameters, which makes it suitable for potential edge deployment~\cite{shuvo2022efficient}. Moreover, this model can be integrated into continuous glucose monitoring devices for real-time decision support.

\begin{figure}[h]
\vspace{-2mm}
    \centering
    \includegraphics[scale=0.3]{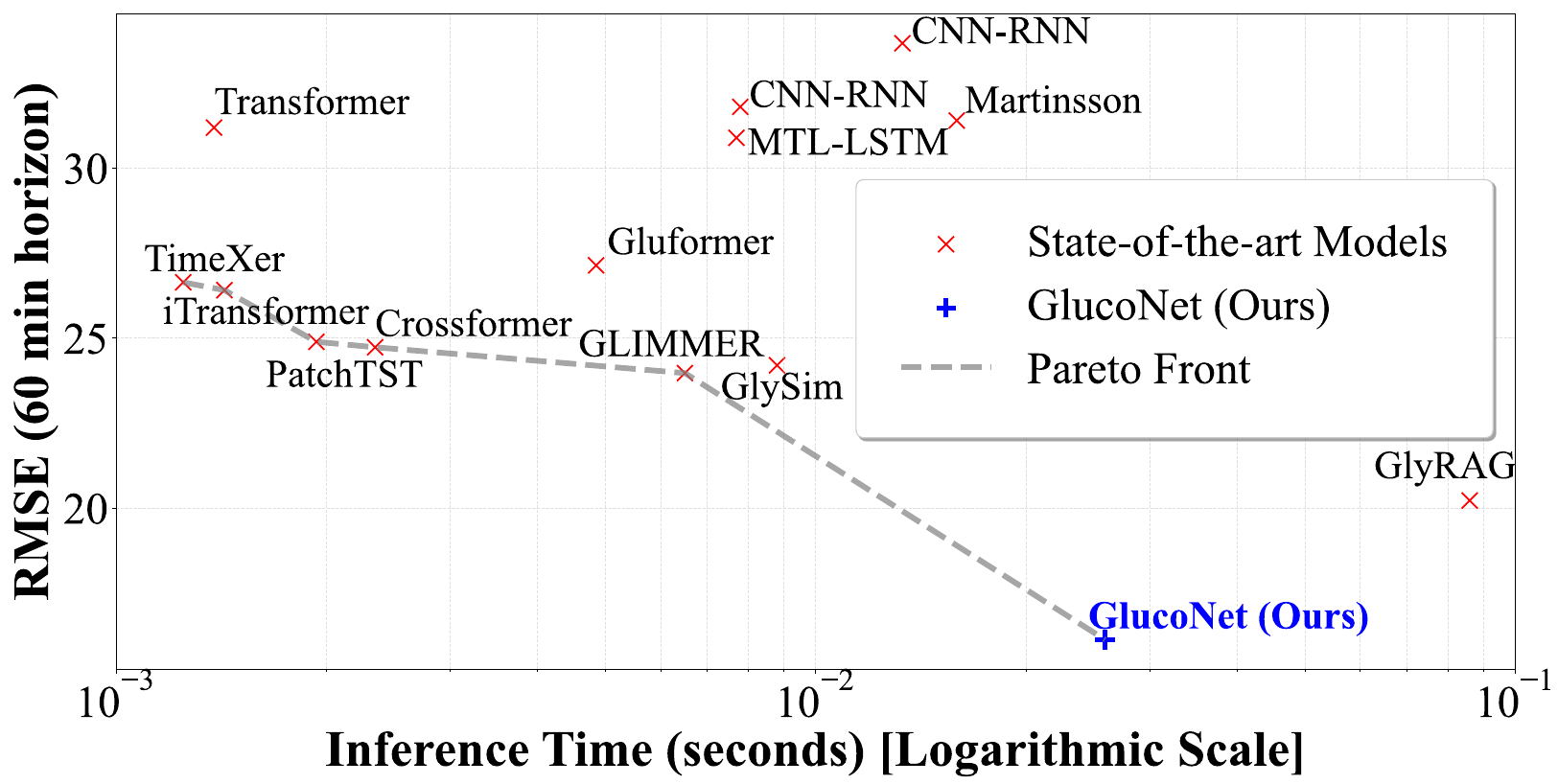}
    \vspace{-2mm}
    \caption{\small Accuracy-efficiency trade-off analysis comparing RMSE and real-world inference latency on a Raspberry Pi 5. The dashed line represents the Pareto front, illustrating that GlucoNet (GlucoNet $+$ KD $(ST)$ $\{[(32,16)]\}$ ) achieves superior predictive performance while maintaining computational efficiency suitable for edge deployment.}
    \label{fig:delay-eff trade}
    \vspace{-2mm}
\end{figure}

To further evaluate the practical deployment feasibility of GlucoNet for real-time mobile health applications, we benchmarked the real-world inference latency of our proposed model against existing state-of-the-art (SOTA) methods on a Raspberry Pi 5. As illustrated in the design space of Fig.~\ref{fig:delay-eff trade}, while a configuration of GlucoNet (GlucoNet $+$ KD $(ST)$ $\{[(32,16)]\}$ ) exhibits a higher inference runtime (0.0259 s) compared to the models such as TimeXer (0.0012 s), it achieves a substantially lower RMSE for the 60-minute PH.

In the clinical context of blood glucose monitoring, where physiological changes typically manifest over 15–60 minutes and sensors sample data at 5-minute intervals, a sub-second inference delay is negligible. Consequently, GlucoNet establishes a new Pareto-optimal frontier. It provides the highest accuracy reported to date within a latency range compatible with edge-integrated CGM systems. This balance confirms that GlucoNet is not only theoretically superior in accuracy but also computationally efficient enough for real-time decision support on resource-constrained wearable devices.

\subsection{\textcolor{black}{Ablation Study}}
We conduct an ablation study to systematically evaluate the impact of architectural components, input modalities, and prediction horizons on GlucoNet’s forecasting performance.

\subsubsection{Effect of Model Architecture Across Prediction Horizons}
We evaluate GlucoNet across three forecasting horizons: short-term (5 minutes), medium-term (30 minutes), and long-term (60 minutes). Table~\ref{tab:avg_ph_result} reports the average error metrics for different GlucoNet configurations compared with the different configuration of GlucoNet across all PHs. The configurations of GlucoNet that we considered are GlucoNet with or without knowledge distillation (KD) and implemented with the large transformer (teacher transformer) or the small transformer (student transformer). These configurations are GlucoNet  $(LT)$ $\{[(128,64)]\}$, GlucoNet $(ST)$ $\{[(128,64)]\}$, and GlucoNet $+$ KD $(ST)$ $\{[(128,64)]\}$ . 

\begin{table*}[t]
\centering
\caption{\small Average error metrics for different configurations of GlucoNet with or without KD across three Prediction Horizons (5 minutes, 30 minutes, 60 minutes). PH: Prediction Horizon, KD: Knowledge Distillation. Parameters: GlucoNet$(LT)\{(128,64)\}$, GlucoNet $(ST)\{(128,64)\}$, GlucoNet$+$KD$(ST)\{(128,64)\}$.}
\setlength{\tabcolsep}{1.3pt}
\small     
\scalebox{0.86}{
\begin{tabular}{c|ccc|ccc|ccc|ccc|ccc|ccc|ccc|ccc|ccc}
\toprule
& \multicolumn{9}{c|}{\textbf{RMSE}} & \multicolumn{9}{c|}{\textbf{MAE}} & \multicolumn{9}{c}{\textbf{$R^2$ Square}} \\
\cmidrule(l){2-28}
{}& \multicolumn{3}{c|}{Total} & \multicolumn{3}{c|}{2018} & \multicolumn{3}{c|}{2020} & \multicolumn{3}{c|}{Total} & \multicolumn{3}{c|}{2018} & \multicolumn{3}{c|}{2020} & \multicolumn{3}{c|}{Total} & \multicolumn{3}{c|}{2018} & \multicolumn{3}{c}{2020} \\ 

PH& 5  & 30  & 60  & 5  & 30  & 60  & 5  & 30  & 60  & 5  & 30  & 60  & 5  & 30  & 60  & 5  & 30  & 60  & 5  & 30  & 60  & 5  & 30  & 60  & 5  & 30  & 60  \\ 
\midrule

Baseline & 8.03 & 14.17 & 16.46 & 8.41 & 15.35 & 16.91 & 7.66 & 13.00 & \textcolor{black}{16.01} & 5.07 & 9.69 & 10.92 & 5.19 & 10.53 & 11.47 & 4.94 & 8.84 & \textcolor{black}{10.37} & 0.97 & 0.91 & 0.89 & 0.96 & 0.89 & 0.86 & 0.98 & 0.94 & 0.92 \\ 
\midrule
\midrule

\makecell{GlucoNet$(LT)$} & 6.31 & 10.18 & 16.30 & 7.61 & 8.98 & \textcolor{black}{13.97} & \textcolor{black}{5.00} & 11.38 & 18.62 & 3.50 & 7.18 & 10.73 & 4.17 & 6.53 & \textcolor{black}{9.67} & \textcolor{black}{2.84} & 7.82 & 11.80 & 0.98 & 0.91 & 0.9 & 0.98 & 0.94 & 0.93 & 0.99 & 0.89 & 0.86 \\ 
\hline

\makecell{GlucoNet $(ST)$} & 6.08 & 10.38 & 16.79 & 7.03 & 8.96 & 14.24 & 5.13 & 11.79 & 19.34 & 3.42 & 7.21 & 11.01 & 3.90 & \textcolor{black}{6.41} & 9.70 & 2.95 & 8.01 & 12.33 & 0.98 & 0.9 & 0.88 & 0.98 & 0.94 & 0.93 & 0.99 & 0.85 & 0.83 \\ 
\hline

\makecell{GlucoNet$+$KD\\$(ST)$}& \textcolor{black}{5.89} & \textcolor{black}{10.03} & \textcolor{black}{16.13} & \textcolor{black}{6.51} & \textcolor{black}{9.07} & 14.37 & 5.28 & \textcolor{black}{10.99} & 17.89 & \textcolor{black}{3.34} & \textcolor{black}{6.95} & \textcolor{black}{10.67} & \textcolor{black}{3.79} & 6.52 & 9.96 & 2.89 & \textcolor{black}{7.38} & 11.39 & 0.99 & 0.91 & 0.89 & 0.98 & 0.94 & 0.93 & 0.99 & 0.89 & 0.86 \\ 
\bottomrule
\end{tabular}
}
\label{tab:avg_ph_result}
\end{table*}

Table~\ref{tab:avg_ph_result} shows that GlucoNet deployed with knowledge distillation achieves efficient models without compromising performance metrics compared to GlucoNet without KD and implemented with the large Transformer.

\begin{figure*}[t]
    \centering
    \includegraphics[scale=0.45]{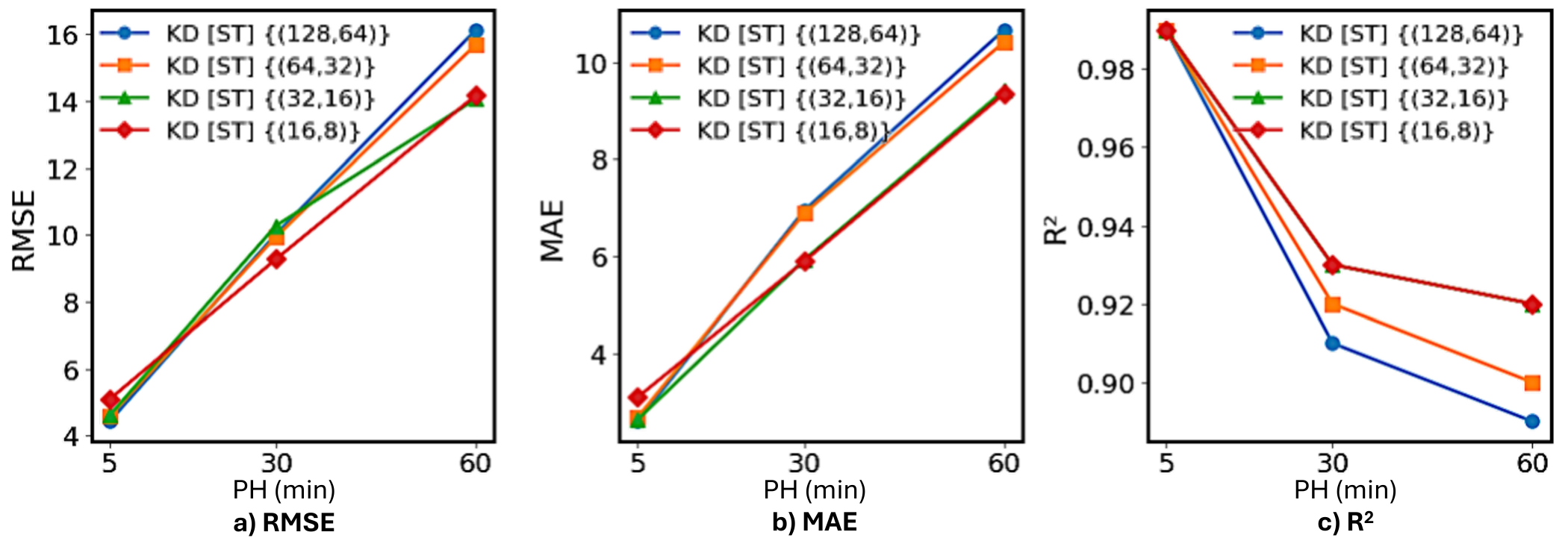}
    \caption{Average error metrics such as a) RMSE, b) MAE, and c) $R^2$  for different configurations of GlucoNet for different memory cells of LSTM across three PHs (5 minutes, 30 minutes, 60 minutes). KD: Knowledge Distillation.}
    \label{fig:avg_total_result}
    \vspace{-3mm}
\end{figure*}

Moreover, the average values of RMSE, MAE, and $R^2$ Square values across different prediction horizons (5 minutes, 30 minutes, and 60 minutes) for multiple GlucoNet configurations are presented in Fig.~\ref{fig:avg_total_result}. Each configuration corresponds to an LSTM with a different number of memory units. The configurations that are considered for comparison are GlucoNet $+$ KD $(ST)$ $\{[(128,64)]\}$, \}, GlucoNet $+$ KD $(ST)$ $\{[(64,32)]\}$,GlucoNet $+$ KD $(ST)$ $\{[(32,16)]\}$, and GlucoNet $+$ KD $(ST)$ $\{[(16,8)]\}$. 

As observed in Fig.~\ref{fig:avg_total_result}, the GlucoNet $+$ KD $(ST)$ $\{[(64,32)]\}$ configuration consistently achieves the best overall accuracy across prediction horizons. Increasing the number of LSTM units or layers beyond this configuration does not yield additional performance gains and may instead lead to overfitting, resulting in reduced predictive accuracy.

\subsubsection{Contribution of Architectural Components and Input Features:}
To identify the sources of GlucoNet’s performance gains, we conducted an ablation study that systematically evaluates the impact of architectural components and input-feature combinations. Specifically, we compare five model configurations—KD (full GlucoNet), LSTM/Small, LSTM/Large, Transformer without VMD, and LSTM without VMD—under four input settings: CGM only, CGM + insulin, CGM + carbohydrates, and all inputs.This ablation isolates the relative contributions of signal decomposition, model architecture, and input modality to GlucoNet’s predictive performance.

\begin{figure}[h]
\vspace{-2mm}
    \centering
    \includegraphics[scale=0.265]{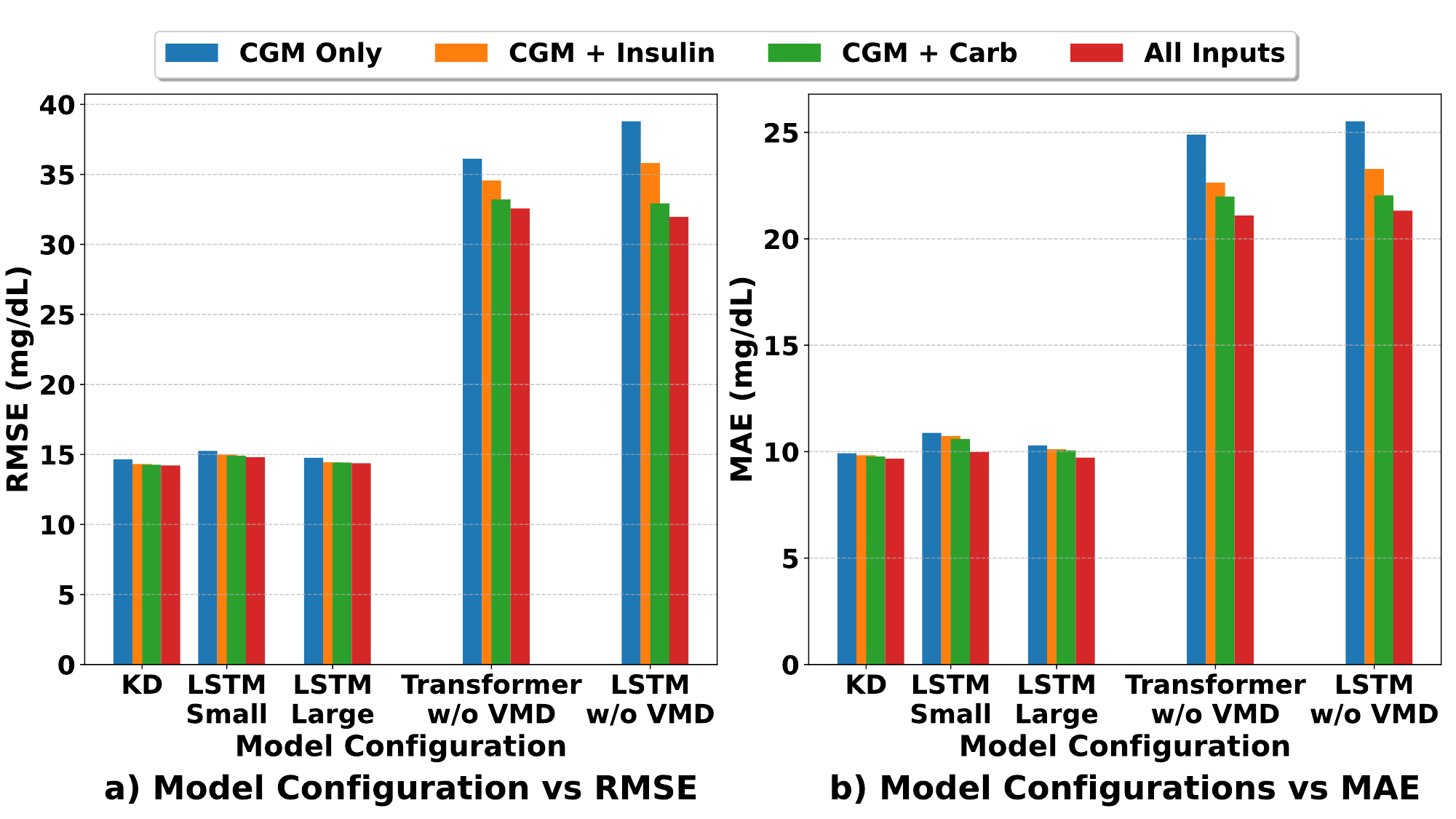}
    \caption{\small \textcolor{black}{Ablation results comparing a) RMSE and b) MAE across model architectures and input-feature combinations. VMD-based parallel models consistently outperform architectures without signal decomposition, regardless of input modality.}}
    \label{fig:ablation_study}
    \vspace{-5mm}
\end{figure}

The results shown in Fig.~\ref{fig:ablation_study} indicate that architectures incorporating Variational Mode Decomposition (VMD) consistently achieve substantially lower error than models operating directly on raw CGM signals. Removing the VMD block results in a marked increase in error, with LSTM-based models exhibiting RMSE values exceeding $32 mg/dL$ even when full multimodal inputs are provided. This indicates that decomposing the non-stationary glucose signal into low- and high-frequency components is critical for stable and accurate forecasting.

\textcolor{black}{While incorporating carbohydrate and insulin information reduces prediction error across all configurations, architectural design plays a dominant role. Models with compact LSTM or KD-based architectures using CGM alone outperform simpler models lacking VMD despite the latter having access to richer input features. The Transformer encoder contributes additional gains only when paired with VMD, where it captures rapid, high-frequency fluctuations that are not effectively modeled by recurrent layers.
Overall, the ablation study confirms that GlucoNet’s accuracy improvements stem primarily from its decomposition-driven parallel architecture, with multimodal inputs providing complementary refinements rather than being the principal source of performance gain.}

\subsubsection{\textcolor{black}{VMD Frequency Decomposition Analysis}}
\textcolor{black}{To clarify the frequency separation process used in Variational Mode Decomposition (VMD), we conducted a systematic design space exploration over the number of intrinsic mode functions (IMFs) and the cutoff index used to group IMFs into low- and high-frequency components. The cutoff index defines the maximum IMF assigned to the low-frequency group, with remaining higher-index modes aggregated as the high-frequency component.}

\begin{figure}[h]
\vspace{-2mm}
    \centering
    \includegraphics[scale=0.265]{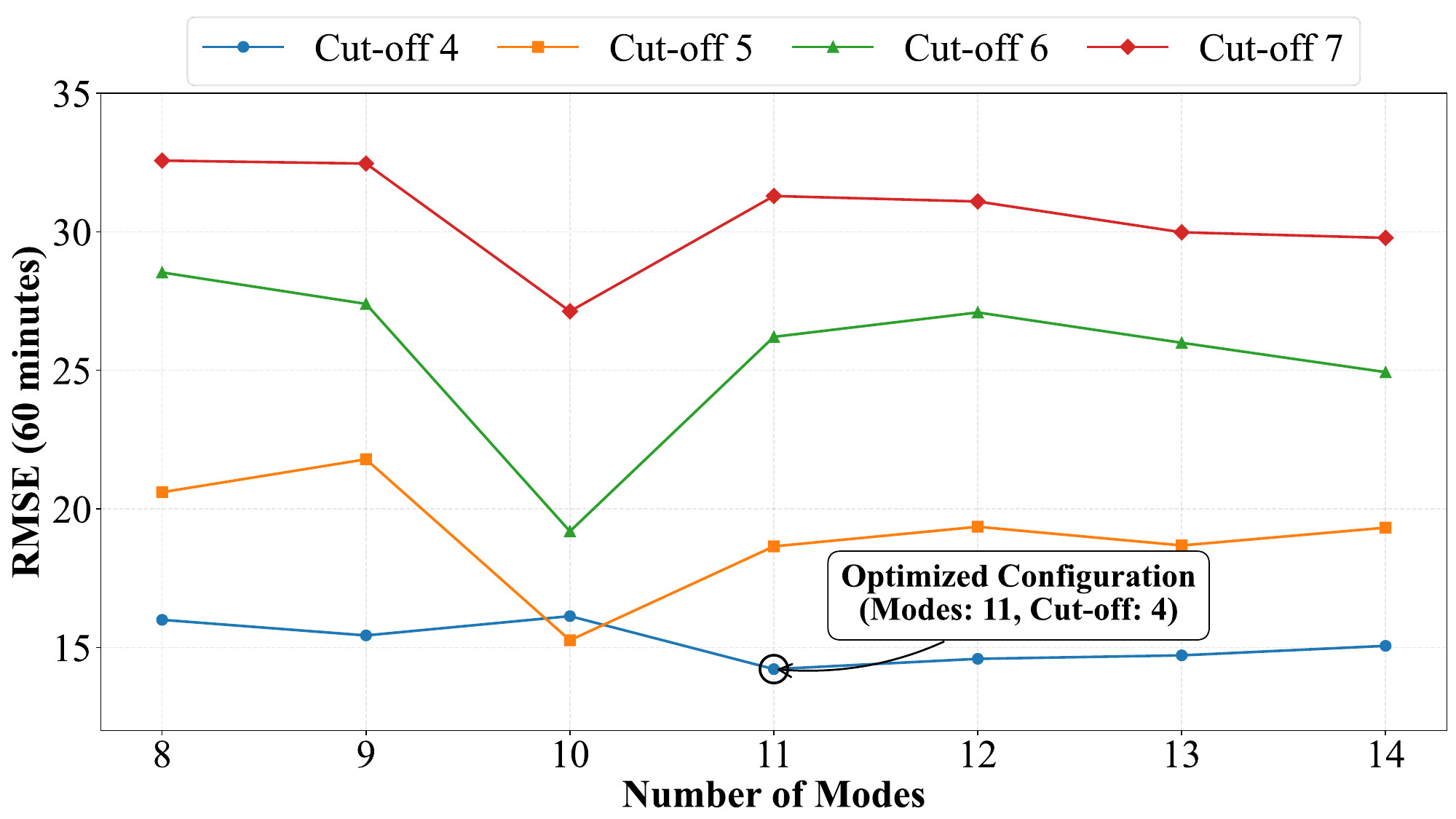}
    \vspace{-2mm}
    \caption{\small \textcolor{black}{Design space exploration of Variational Mode Decomposition (VMD) parameters. RMSE for the 60-minute prediction horizon is reported across varying numbers of intrinsic mode functions (IMFs) and cutoff indices used to separate low- and high-frequency components. The optimal configuration (11 IMFs, cutoff = 4) achieves the lowest long-horizon prediction error and is selected for GlucoNet.}}
    \label{DS_VMD}
    \vspace{-5mm}
\end{figure}

\textcolor{black}{We evaluated configurations with 8 to 14 IMFs and cutoff indices ranging from 4 to 7. For each configuration, forecasting performance was assessed using RMSE and MAE across 30-minute and 60-minute prediction horizons. Fig.}~\ref{DS_VMD} \textcolor{black}{summarizes the resulting performance landscape for the 60-minute horizon, illustrating the strong interaction between the number of modes and the cutoff selection.}

\textcolor{black}{The results indicate that assigning too many modes to the high-frequency group leads to unstable long-horizon predictions, while overly large cutoff values oversmooth transient glucose dynamics. The optimal configuration is achieved with 11 IMFs and a cutoff index of 4, which yields the lowest prediction error for long-horizon forecasting. This configuration is therefore adopted in the final GlucoNet architecture.}

\begin{table}[]
\textcolor{black}{
\centering
\caption{\textcolor{black}{Clinical evaluation of hypoglycemic (hypo) and hyperglycemic (hyper) events for different configurations of GlucoNet.}}
\label{tab:clinical_results}
\footnotesize
\begin{tabular}{lcc|cc|cc}
\toprule
\textbf{Dataset} &
\multicolumn{2}{c|}{\textbf{Accuracy}} &
\multicolumn{2}{c|}{\textbf{RMSE}} &
\multicolumn{2}{c}{\textbf{Samples}} \\
\cline{2-7}  
\textbf{(Method)} & \textbf{Hypo} & \textbf{Hyper} & \textbf{Hypo} & \textbf{Hyper} & \textbf{Hypo} & \textbf{Hyper} \\
\hline
Ohio-LT & 0.715 & 0.915 & 14.655 & 15.030 & \multirow{3}{*}{143}& \multirow{3}{*}{1020} \\
Ohio-ST & 0.595 & 0.920 & 13.805 & 15.300  \\
\kdboldred{Ohio-KD} & \kdboldred{0.530} & \kdboldred{0.925} & \kdboldred{13.435} & \kdboldred{14.285} \\
\midrule
AZ-LT   & 0.515 & 0.855 & 18.765 & 15.515 & \multirow{3}{*}{72}  & \multirow{3}{*}{528} \\
AZ-ST   & 0.515 & 0.860 & 18.735 & 14.915 &  \\
\kdboldred{AZ-KD}   & \kdboldred{0.405} & \kdboldred{0.840} & \kdboldred{18.895} & \kdboldred{14.740} \\
\bottomrule
\end{tabular}
}
\vspace{-2mm}
\end{table}
\begin{table*}[ht]
\textcolor{black}{
\centering
\caption{\textcolor{black}{Performance comparison with Pearson correlation ($R^2$) across age groups and gender for hypoglycemic (Hypo) and hyperglycemic (Hyper) events. Ohio results are shown on the first row, with corresponding AZT1D results shown below in parentheses.}}
\label{tab:age_gender_results}
\footnotesize
\begin{tabular}{clcc|cc|cc|cc|cc|cc|c}
\toprule
\multicolumn{2}{c}{\textbf{Group}} & \multicolumn{2}{c|}{\textbf{Accuracy}} & \multicolumn{2}{c|}{\textbf{RMSE}} & \multicolumn{2}{c|}{\textbf{Samples}} & \multicolumn{2}{c|}{\textbf{PH=5min}} & \multicolumn{2}{c|}{\textbf{PH=30min}} & \multicolumn{2}{c|}{\textbf{PH=60min}} & \multirow{2}{*}{\textbf{$R^2$}} \\
\cline{3-14}
& & \textbf{Hypo} & \textbf{Hyper} & \textbf{Hypo} & \textbf{Hyper} & \textbf{Hypo} & \textbf{Hyper} & \textbf{RMSE} & \textbf{MAE} & \textbf{RMSE} & \textbf{MAE} & \textbf{RMSE} & \textbf{MAE} & \\
\midrule
\multirow{6}{*}{\rotatebox{90}{\textbf{Age}}} 
& \multirow{2}{*}{20--40} & 0.64 & 0.90 & 12.65 & 16.28 & 220 & 670 & 6.22 & 14.16 & 6.47 & 13.91 & 8.18 & 17.19 & 0.90 \\
& & (0.26) & (0.75) & (19.13) & (13.65) & (108) & (326) & (6.55) & (10.44) & (6.12) & (10.43) & (6.85) & (12.73) & (0.87) \\
\cline{2-15}
& \multirow{2}{*}{41--60} & 0.44 & 0.94 & 14.05 & 12.88 & 103 & 1181 & 5.63 & 12.56 & 5.99 & 12.41 & 7.48 & 15.17 & 0.93 \\
& & (0.60) & (0.89) & (13.45) & (14.37) & (66) & (623) & (5.86) & (12.29) & (6.13) & (12.01) & (7.44) & (14.56) & (0.87) \\
\cline{2-15}
& \multirow{2}{*}{$>$60} & 0.68 & 0.86 & 9.30 & 17.32 & 161 & 749 & 7.63 & 17.80 & 7.77 & 17.17 & 9.65 & 20.99 & 0.82 \\
& & (0.33) & (0.84) & (21.78) & (15.22) & (67) & (536) & (7.36) & (13.55) & (7.14) & (13.25) & (8.41) & (15.75) & (0.87) \\
\midrule
\multirow{4}{*}{\rotatebox{90}{\textbf{Gender}}} 
& \multirow{2}{*}{Male} & 0.43 & 0.92 & 12.64 & 13.16 & 110 & 1003 & 5.86 & 12.79 & 6.17 & 12.49 & 7.67 & 15.19 & 0.92 \\
& & (0.40) & (0.83) & (18.82) & (13.44) & (73) & (447) & (6.85) & (12.25) & (6.64) & (11.83) & (7.74) & (14.19) & (0.88) \\
\cline{2-15}
& \multirow{2}{*}{Female} & 0.63 & 0.93 & 14.23 & 15.41 & 175 & 1037 & 6.06 & 14.24 & 6.38 & 14.15 & 8.07 & 17.53 & 0.91 \\
& & (0.41) & (0.85) & (18.97) & (16.04) & (71) & (609) & (6.80) & (13.26) & (6.78) & (13.13) & (8.09) & (15.73) & (0.86) \\
\bottomrule
\end{tabular}
}
\vspace{-3mm}
\end{table*}

\subsection{Clinical Robustness and Generalizability Analysis}
Beyond average forecasting accuracy, it is critical to evaluate model behavior under clinically significant conditions and across diverse patient populations. This section examines the robustness of GlucoNet with respect to abnormal glycemic events, demographic variability, and cross-dataset generalization.
\subsubsection{Event-Based and Demographic Robustness} 

To evaluate the practical reliability of GlucoNet, we performed a granular analysis of its performance during critical glycemic events and across diverse demographic groups. 
Table~\ref{tab:clinical_results} compares the three GlucoNet configurations and shows that the KD variant consistently provides the best overall trade‑off between clinical accuracy and model complexity across both cohorts. In particular, KD achieves the lowest RMSE for hyperglycemic events (14.3mg/dL on Ohio, 14.7mg/dL on AZT1D) while maintaining comparable hypoglycemia errors, despite the strong class imbalance where hyper events are far more frequent than hypo events (e.g., 1020 vs. 143 events in Ohio). These results motivate using GlucoNet‑KD as the reference configuration for subsequent clinical analysis.

\textcolor{black}{
Furthermore, Table~\ref{tab:age_gender_results} details the performance of the GlucoNet-KD model across different age and gender cohorts to ensure demographic equity. 
Across prediction horizons, accuracy and Pearson correlation coefficients ($R^2$) are similar for adults aged 20–40 and 41–60, whereas participants $>$60 years exhibit noticeably lower $R^2$ and higher long‑horizon RMSE, indicating reduced robustness in older adults who are under‑represented and likely more heterogeneous clinically. Gender effects are comparatively modest: males and females show similar hyper/hypo detection accuracy and RMSE, with only small and dataset‑specific differences, suggesting that GlucoNet‑KD generalizes similarly across genders while age, especially $>$60 years, emerges as the more influential clinical factor.
These results indicate that by integrating multimodal inputs of insulin and carbohydrates with signal decomposition, GlucoNet provides a generalized solution that adapts to varying metabolic profiles across the lifespan.}

\subsubsection{\textcolor{black}{Cross-Dataset Generalization}}

\textcolor{black}{To further evaluate the robustness and generalizability of GlucoNet, we conducted two experiments that isolate the effects of carbohydrate intake uncertainty and inter-subject metabolic variability.}

\begin{figure}[h]
    \centering
    \vspace{-2mm}
    \includegraphics[scale=0.265]{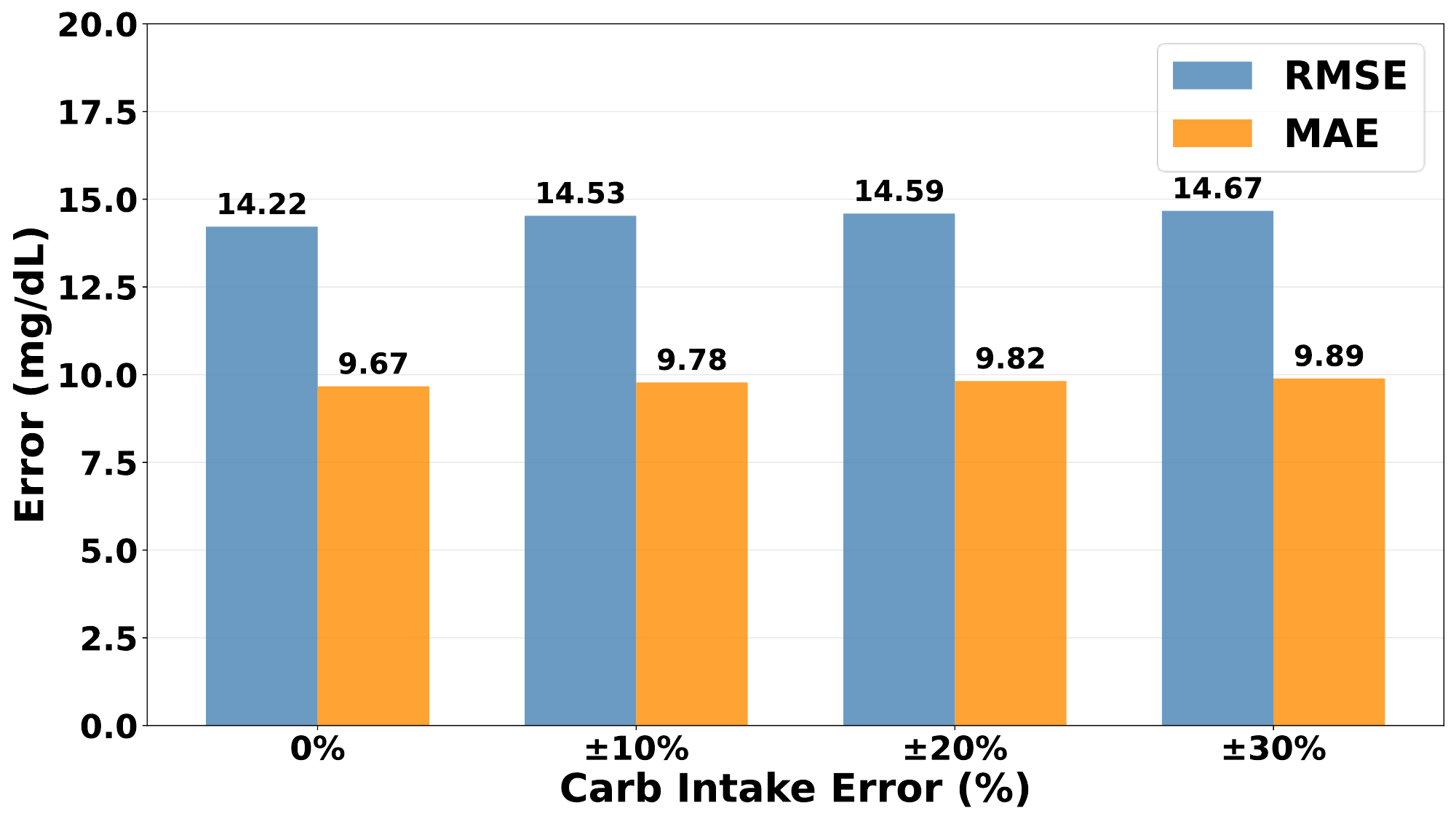}
    \vspace{-2mm}
    \caption{\small \textcolor{black}{Robustness analysis of GlucoNet under controlled carbohydrate intake perturbations. RMSE and MAE are reported as stochastic noise ($\pm10\%$, $\pm20\%$, $\pm30\%$) injected into carbohydrate intake inputs during inference, demonstrating robustness to realistic self-reporting inaccuracies.}}
    \label{fig:ablation_study_Robust}
    \vspace{-2mm}
\end{figure}

\textcolor{black}{In real-world settings, carbohydrate intake information is frequently self-reported and subject to significant inaccuracies. To assess the robustness of GlucoNet under these conditions, we performed a sensitivity analysis by injecting stochastic noise into the carbohydrate intake feature during the inference phase. Noise levels of $\pm10\%$, $\pm20\%$, and $\pm30\%$ were applied to simulate realistic logging errors. As shown in Fig.}~\ref{fig:ablation_study_Robust}, \textcolor{black}{GlucoNet maintains stable predictive performance across all noise levels. Even at $\pm30\%$ deviation, RMSE increases marginally from 14.22 mg/dL to 14.67 mg/dL, with similarly limited variation observed in MAE. These results indicate that GlucoNet integrates carbohydrate intake as a physiologically relevant input while remaining robust to uncertainty in its measurement, by jointly leveraging complementary information across multimodal signals through its decomposition-based parallel architecture.}

\begin{figure}[h]
    \centering
    \vspace{-2mm}
    \includegraphics[scale=0.265]{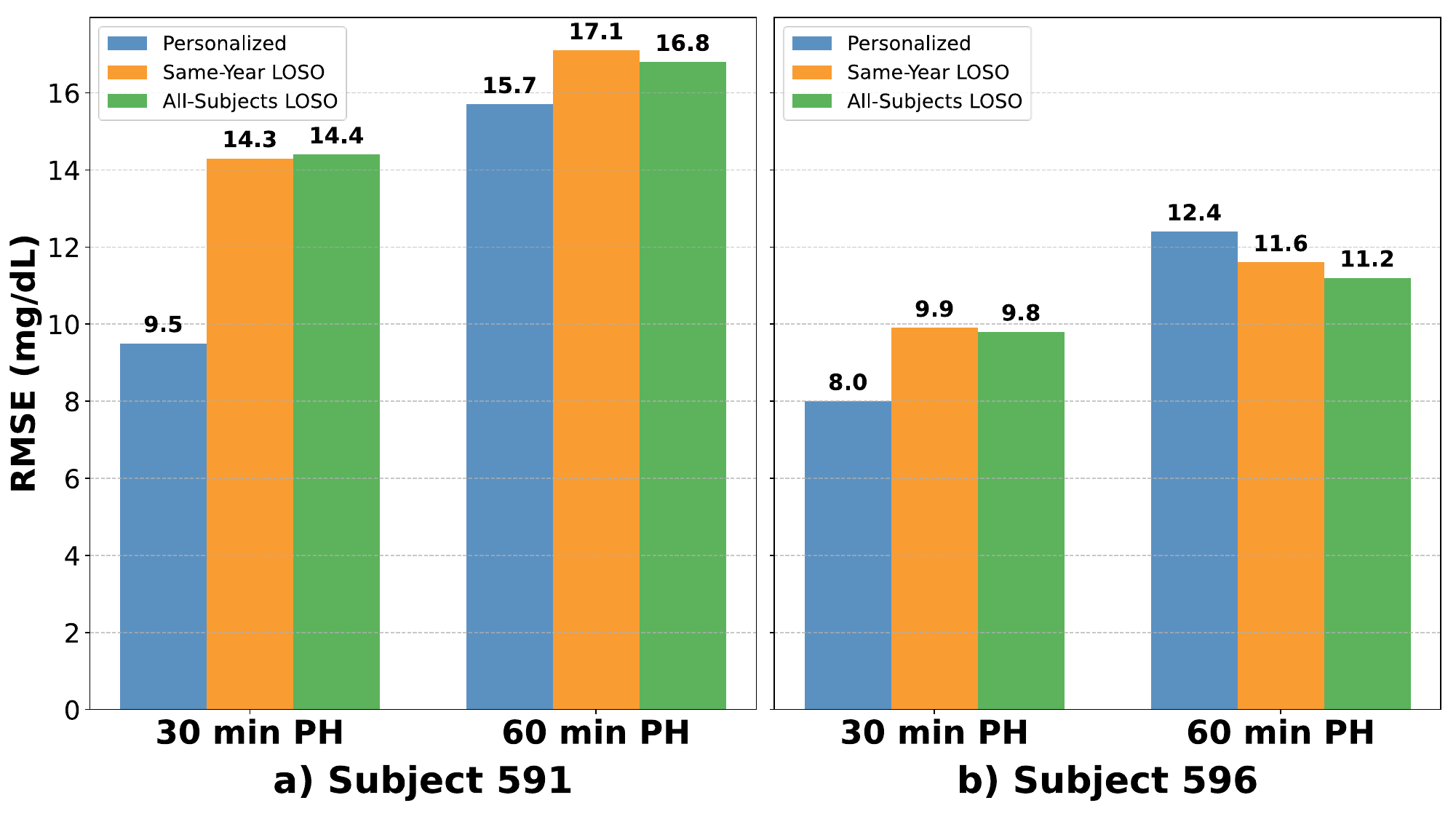}
    \vspace{-2mm}
    \caption{\small \textcolor{black}{Cross-subject generalization performance under Leave-One-Subject-Out (LOSO) evaluation for Subjects a) 591 and b) 596 on the OhioT1DM dataset at 30-minute and 60-minute prediction horizons, comparing personalized, same-year LOSO, and all-subject LOSO training configurations.}}
    \label{fig:ablation_study_LOSO}
    \vspace{-5mm}
\end{figure}

\textcolor{black}{To evaluate robustness to individual metabolic variability, we conducted a Leave-One-Subject-Out (LOSO) cross-validation study on the OhioT1DM dataset. In this setting, the model is trained on data from all but one participant and evaluated on the held-out subject. Fig.}~\ref{fig:ablation_study_LOSO} \textcolor{black}{reports RMSE for representative subjects (591 and 596) under personalized training, same-year LOSO, and all-subject LOSO configurations at 30-minute and 60-minute prediction horizons. While personalized models achieve the lowest error, LOSO configurations preserve competitive accuracy without severe degradation. This stability demonstrates that GlucoNet captures generalized physiological glucose dynamics that transfer across individuals rather than overfitting to subject-specific metabolic patterns.}

\color{black}
\section{Conclusions}
\label{sec:conclusion}
We presented a novel machine learning approach, GlucoNet, for blood glucose forecasting in mobile systems. In designing GlucoNet, we aim to achieve improved accuracy over traditional blood glucose forecasting methods while maintaining computational efficiency. GlucoNet combines blood glucose data with event-based variables such as carbohydrate intake and insulin dosage. To address the non-linear and non-stationary nature of blood glucose time series data, Variational Mode Decomposition (VMD) is employed to decompose the data into low and high-fluctuation signal modes. The time series features of carbohydrate intake and insulin dosage are combined with low and high-fluctuation signal modes and create low-frequency features and high-frequency features. Low-frequency and high-frequency features are then fed to the LSTM and Transformer models to forecast BGL, respectively. To enhance efficiency, a knowledge distillation technique is implemented to compress the Transformer model. \textcolor{black}{ Furthermore, we deliberately adopt a lightweight SSR-based IOB/COB model, which matches the coarse insulin/meal logs in OhioT1DM and AZT1D and captures the delayed onset–peak–decay of their effects while keeping preprocessing simple and latency low for edge deployment. Experimental results show that GlucoNet improves RMSE by approximately $33\%$ while decreasing the total model parameters by $26\%$ compared with existing models for forecasting blood glucose. Evaluation on both the OhioT1DM and AZT1D datasets demonstrates effective generalization to larger and more diverse patient populations. Ablation and robustness analyses confirm that the performance gains are primarily driven by the decomposition-based parallel architecture and remain stable under input uncertainty, demographic variability, and clinically significant glycemic events. Moreover, real-time deployment on a Raspberry Pi shows that GlucoNet achieves a Pareto-optimal accuracy–latency trade-off. These results highlight GlucoNet's potential as a more accurate and efficient tool practical edge-based blood glucose forecasting.}


\bibliographystyle{IEEEtran}
\bibliography{ref}

\vfill

\end{document}